\documentclass[10pt,twocolumn,letterpaper]{article}

\usepackage{cvpr} % uncomment this for the final submission
\usepackage{times}
\usepackage{epsfig}
\usepackage{graphicx}
\usepackage{amsmath}
\usepackage{amssymb}

\usepackage{graphicx}
\usepackage{amsmath}
\usepackage{amssymb}
\usepackage{booktabs}
\usepackage{multirow}
\usepackage{float}
\usepackage{csquotes}

\usepackage{multirow}
\usepackage{algorithm}% http://ctan.org/pkg/algorithm
\usepackage{algpseudocode}% http://ctan.org/pkg/algorithmicx

\begin{document}

%%%%%%%%% TITLE
\title{StableMorph: High-Quality Face Morph Generation with Stable Diffusion}

\author{Wassim Kabbani \quad Kiran Raja \quad Raghavendra Ramachandra \quad Christoph Busch\\
Norwegian University of Science and Technology\\
Gjøvik, Norway\\
{\tt\small \{wassim.h.kabbani; kiran.raja; raghavendra.ramachandra; christoph.busch \} @ntnu.no}
% \and
}

\maketitle
\thispagestyle{empty}

\begin{abstract}

Face morphing attacks threaten the integrity of biometric identity systems by enabling multiple individuals to share a single identity. To develop and evaluate effective morphing attack detection (MAD) systems, we need access to high-quality, realistic morphed images that reflect the challenges posed in real-world scenarios. However, existing morph generation methods often produce images that are blurry, riddled with artifacts, or poorly constructed—making them easy to detect and not representative of the most dangerous attacks. In this work, we introduce StableMorph, a novel approach that generates highly realistic, artifact-free morphed face images using modern diffusion-based image synthesis. Unlike prior methods, StableMorph produces full-head images with sharp details, avoids common visual flaws, and offers unmatched control over visual attributes. Through extensive evaluation, we show that StableMorph images not only rival or exceed the quality of genuine face images, but also maintain a strong ability to fool face recognition systems—posing a greater challenge to existing MAD solutions and setting a new standard for morph quality in research and operational testing. StableMorph improves the evaluation of biometric security by creating more realistic and effective attacks and supports the development of more robust detection systems.

\end{abstract}
\section{Introduction}
\label{sec:intro}

The widespread deployment of face recognition systems (FRS) has made them vulnerable to various kinds of attacks, such as presentation attacks and morphing attacks \cite{Scherhag-MorphingAttacks-Survey-IEEEAccess-2019}. Face image morphing is an image manipulation technique that generates a new image by blending face images from two or more subjects. Given a morphed image, a malicious actor can obtain a fully authentic ID document that can be used by both the applicant themselves and potentially the other individual(s), allowing them to evade identification and gain unauthorized access \cite{Scherhag-MorphingAttacks-MorphingTechniques-BIOSIG-2017}. A good morphed image in a successful attack should be able to deceive a human observer, such as a border guard at a border crossing or an ID document expert at an ID document issuing authority. It should also deceive a FRS and have a high comparison score when the FRS compares the morphed image with a bona fide image of the subject captured as a probe image at an automatic border control gate or taken in a controlled environment to verify the identity of the subject \cite{Venkatesh-FaceMorphingAttackGenerationAndDetection-TTS-2021}.

\begin{figure}[t]
\centering

\begingroup
\setlength{\tabcolsep}{1.3pt} % Default value: 6pt
\begin{tabular}{ccc}
    
    Subject 1 &
    StableMorph &
    Subject 2 \\

    \includegraphics[width=.33\linewidth]{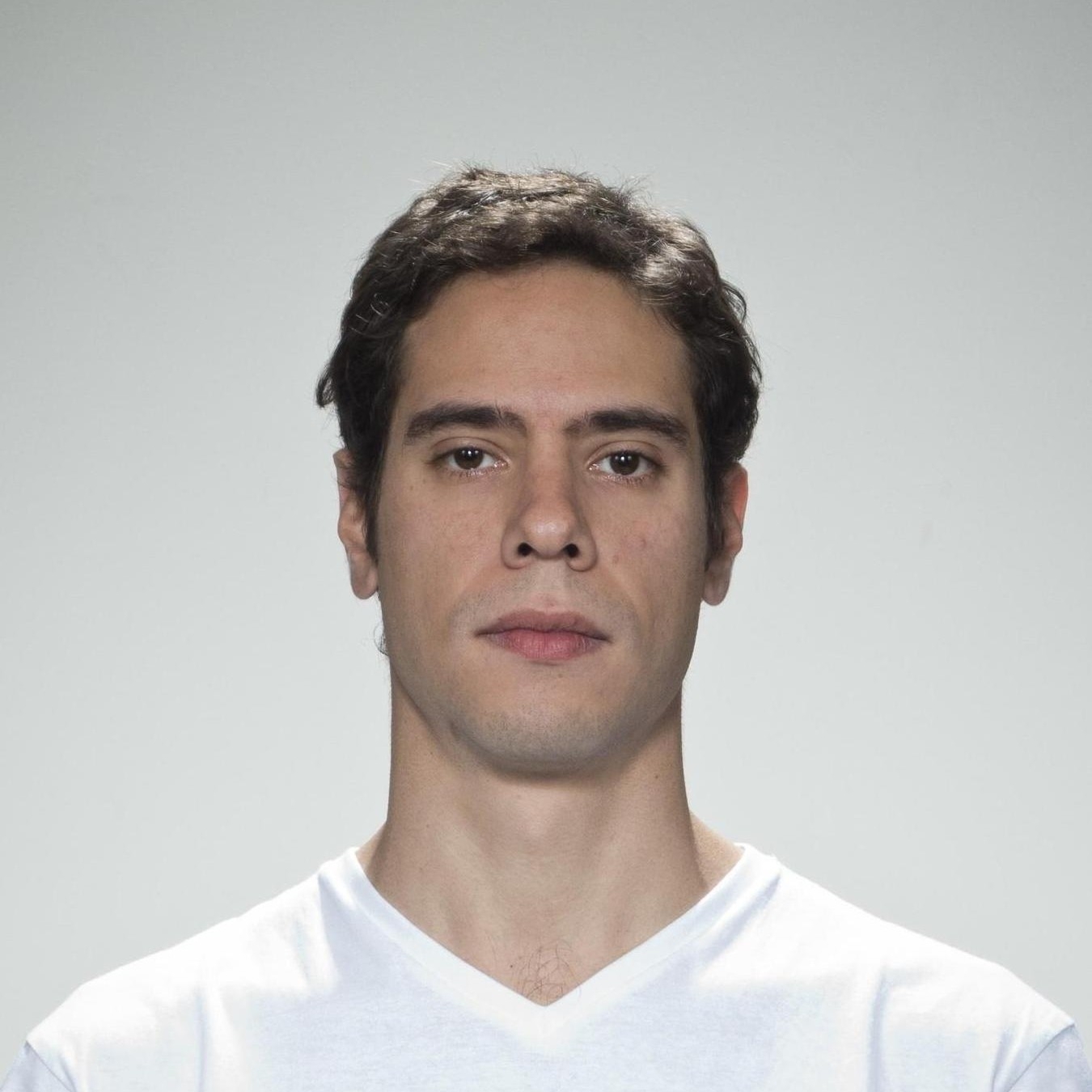} &
    \includegraphics[width=.33\linewidth]{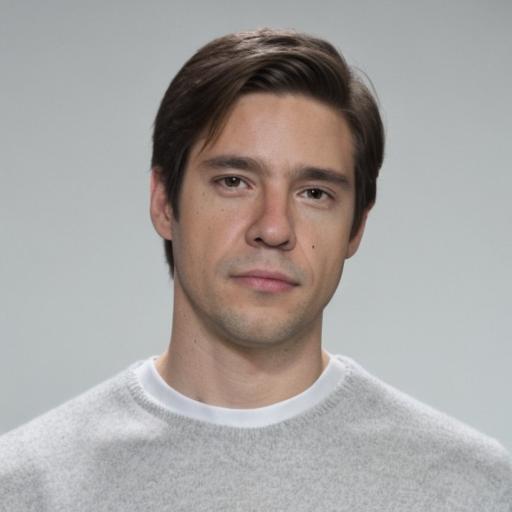} &
    \includegraphics[width=.33\linewidth]{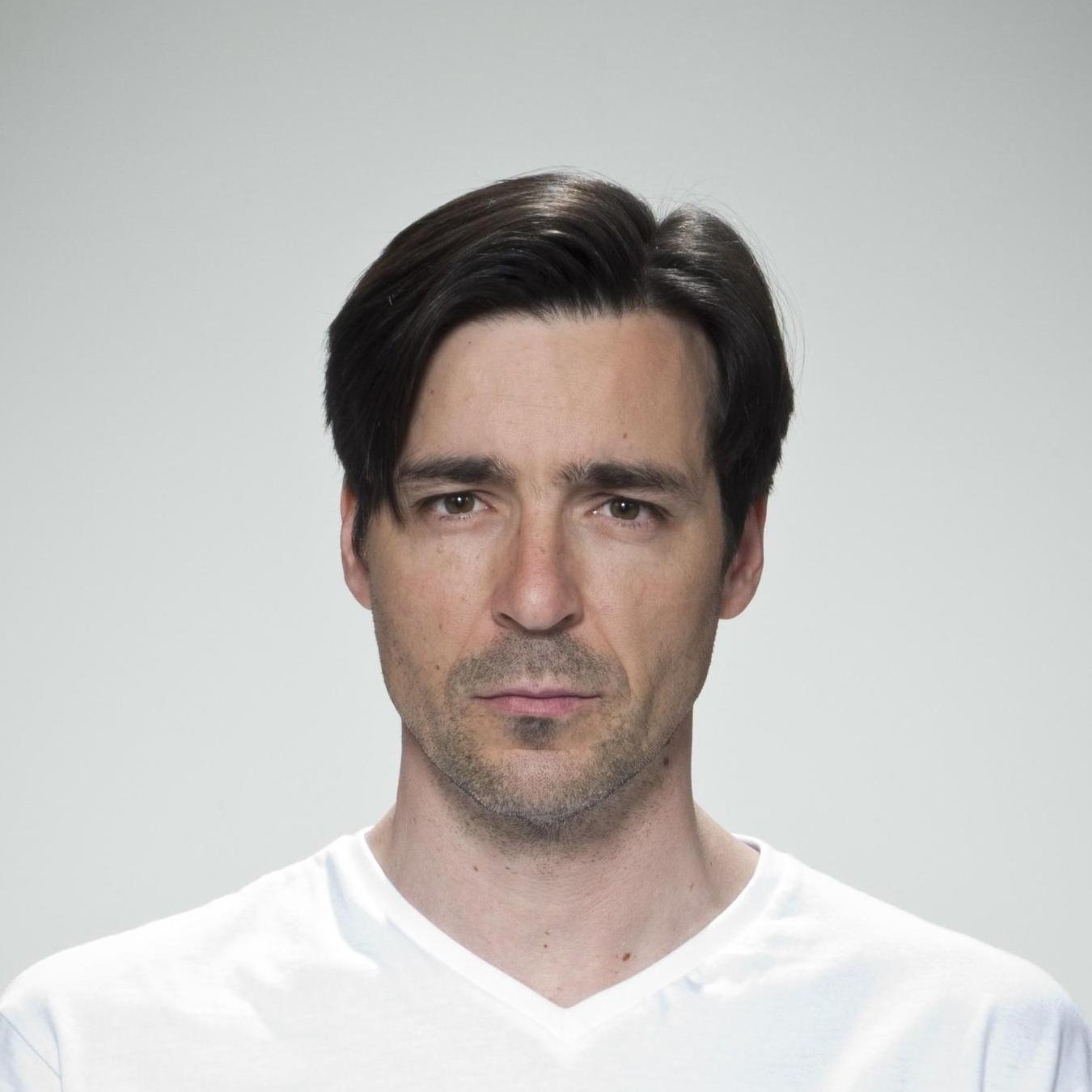} \\

    \includegraphics[width=.33\linewidth]{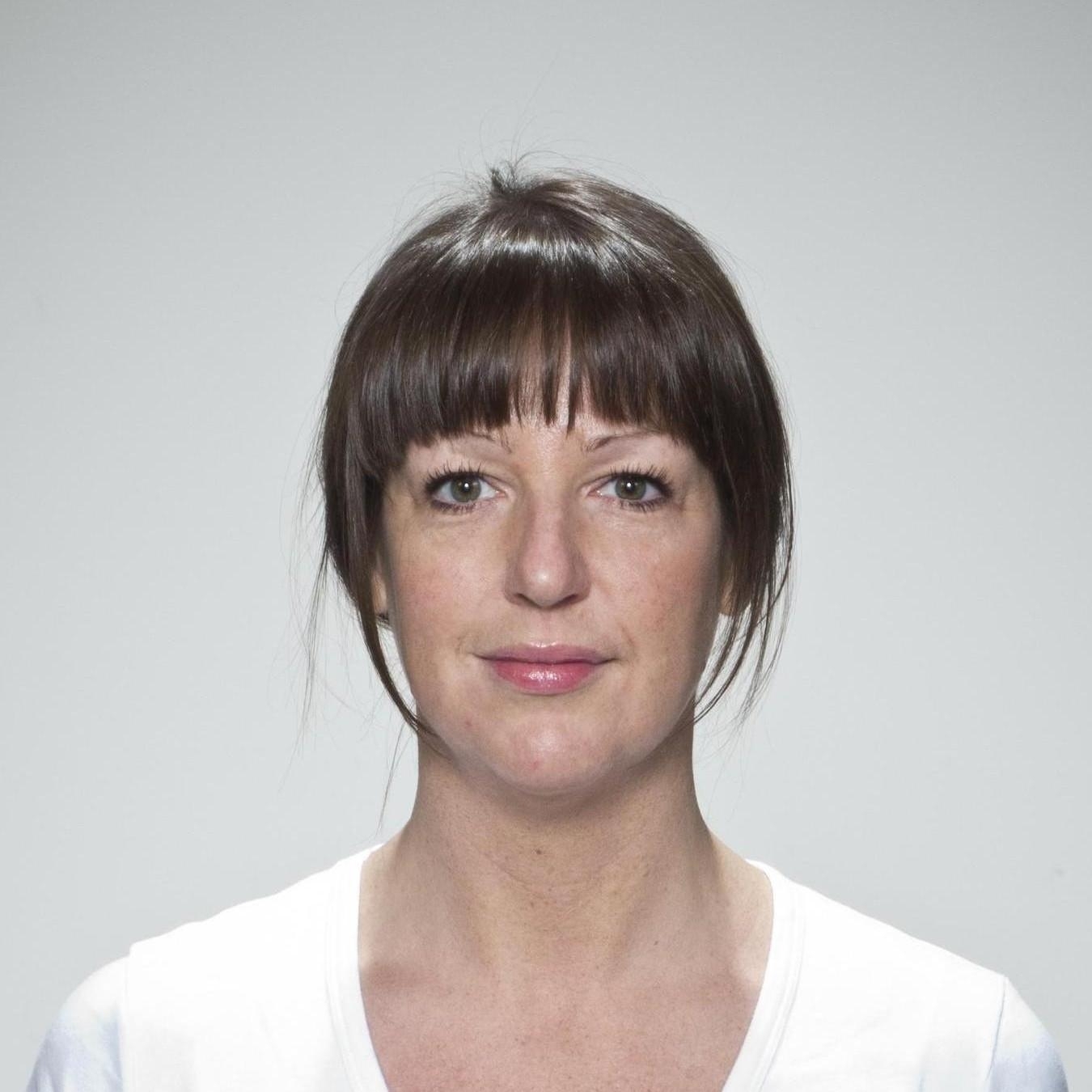} &
    \includegraphics[width=.33\linewidth]{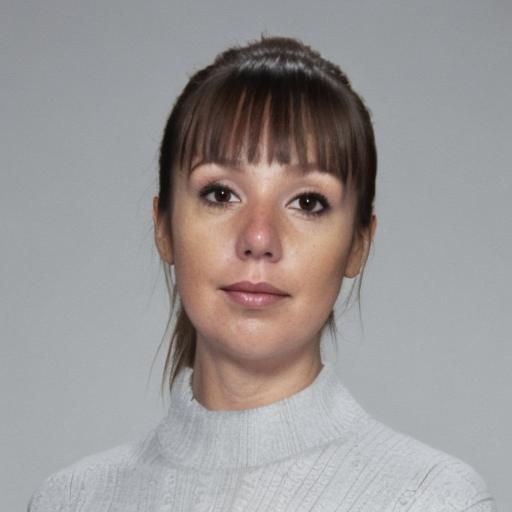} &
    \includegraphics[width=.33\linewidth]{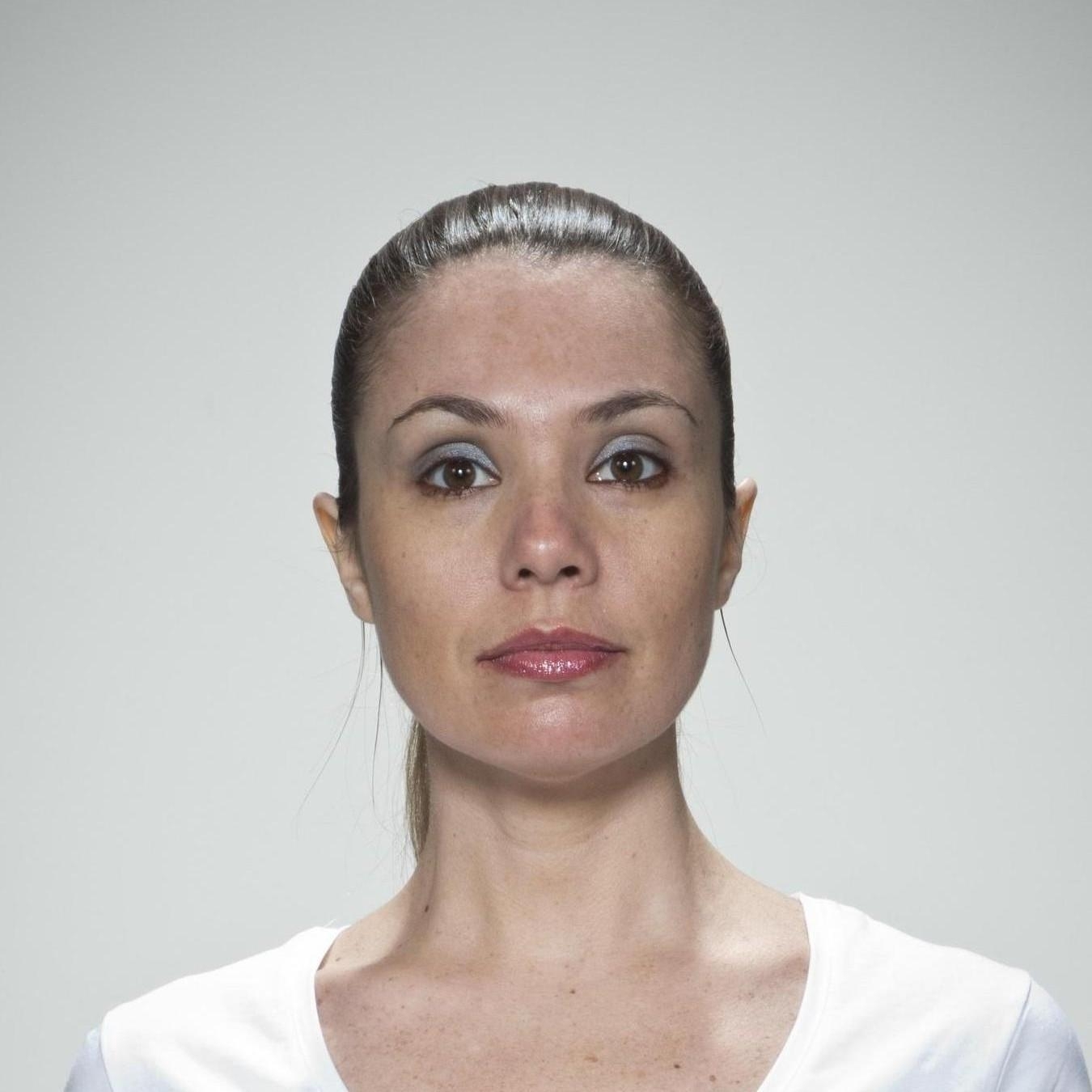} \\
    
    \includegraphics[width=.33\linewidth]{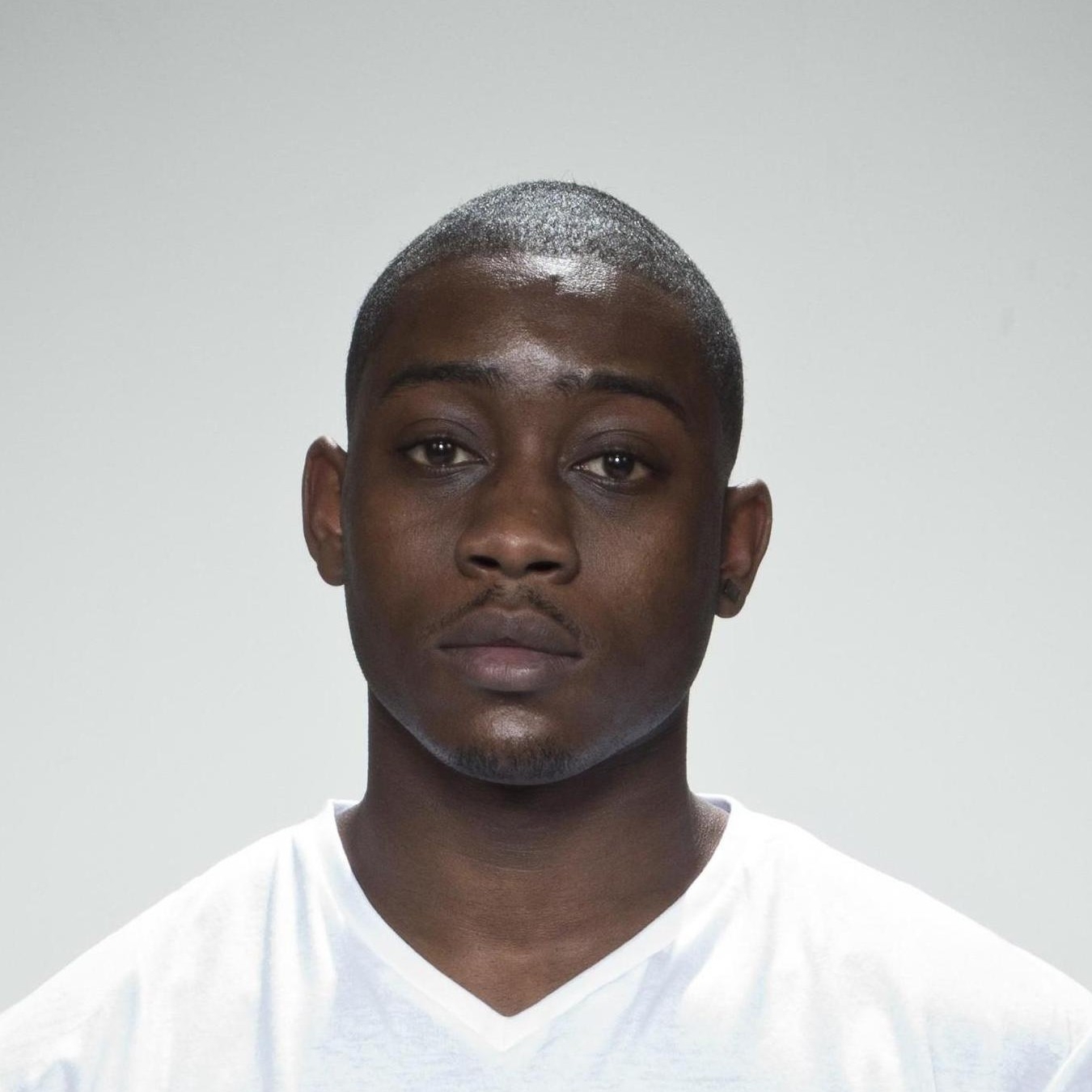} &
    \includegraphics[width=.33\linewidth]{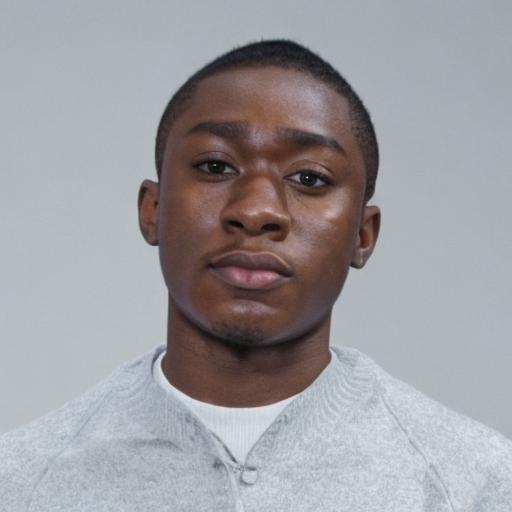} &
    \includegraphics[width=.33\linewidth]{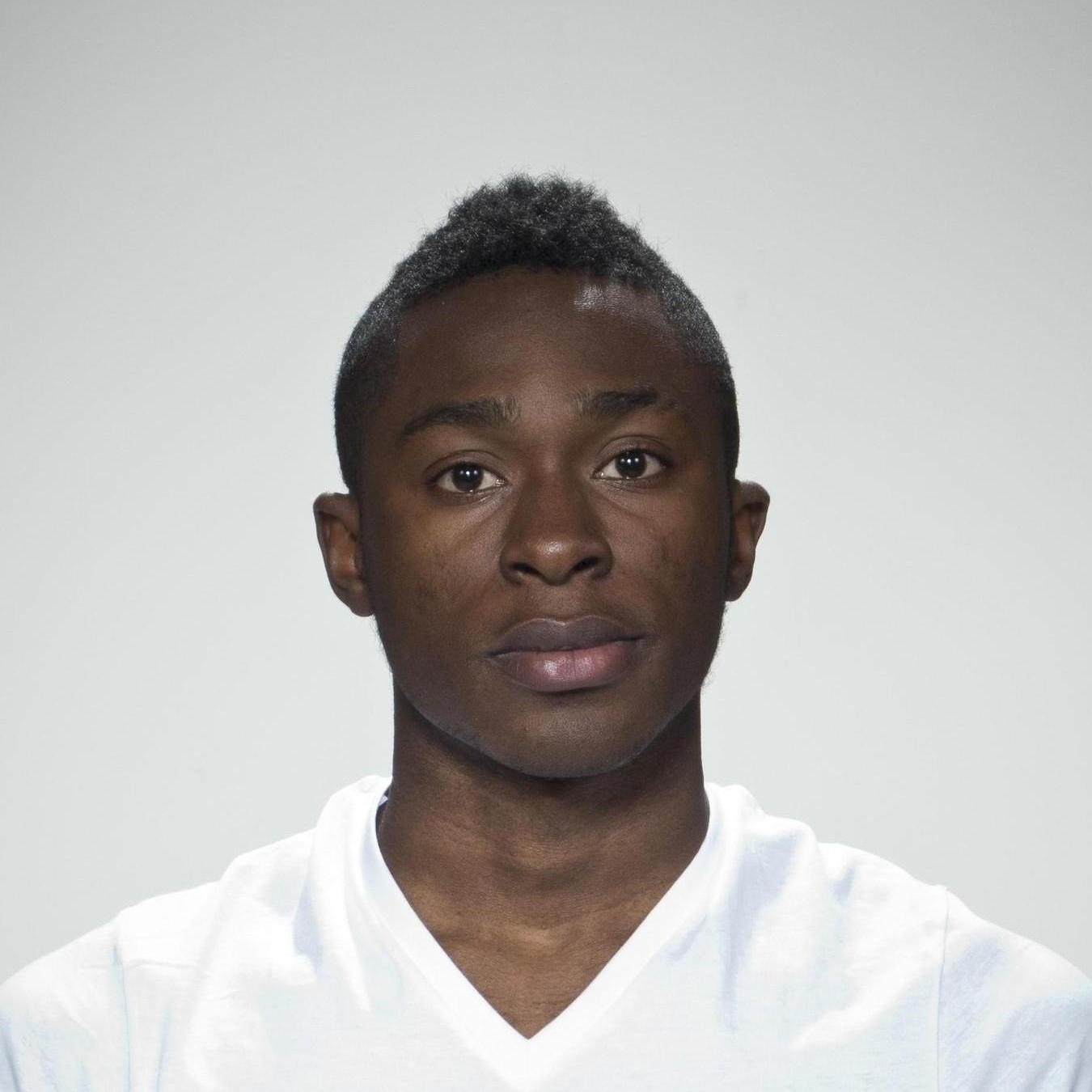} \\

\end{tabular}
\endgroup

\caption{Sample morphed images generated by StableMorph. The 1st and 3rd columns are the bona fide images of the subjects. Images from the FRLL dataset \cite{DeBruine2021}.}
\label{fig:qualitative}
\end{figure}

Existing face morph generation methods can be split into two main categories: landmark-based and deep learning-based \cite{Scherhag-MorphingAttacks-MorphingTechniques-BIOSIG-2017, Venkatesh-FaceMorphingAttackGenerationAndDetection-TTS-2021}. The landmark-based methods identify the face landmark points in input face images and use them to warp the pixels of common facial regions such as nose, eyes, mouth, and eyebrows to more averaged positions and values \cite{Makrushin-FaultlessMorphs-ICVICGTA-2017, Liao2014MorphingSSIM, Ferrara-TextureBlendingAndShapeWarpingInFaceMorphing-IEEE-BIOSIG-2019}. However, these methods are known to create visible artifacts in the morphed images, such as shadow effects around the head, the hair, the nostrils of the nose, and the pupils of the eyes, making them easy to detect by manual or automated inspection. Hence, they are usually followed by some post-processing steps to improve the quality of the image \cite{Venkatesh-FaceMorphingAttackGenerationAndDetection-TTS-2021}. In the methods that try to morph the entire head area, the shadow effects become much more visible around the hair, the ears, and the neck, while the methods that morph only the face area to avoid this problem require the input images to have the faces as perfectly aligned as possible, and since they use the landmarks on the face to replace the original pixels in the face area with the morphed pixels, they become detectable by image forensic tools \cite{Kraetzer2017Media-Forensics, Makrushin2018Mitigating}. The deep learning-based methods, on the other hand, use generative deep learning models, primarily generative adversarial networks (GAN) \cite{Damer-MorGAN-BTAS-2018, venkatesh2020can, Zhang-MIPGAN-MorphingAttacks-IEEE-2021}, to generate the morphed image, incorporating the identities of participating subjects to obtain an image representative of all subjects. Although these methods reduce or avoid the artifacts of the landmark-based methods, the existing methods come with a new set of artifacts, such as images appearing washed out, grainy, and lacking the quality and sharpness of genuine images \cite{Zhang-MIPGAN-MorphingAttacks-IEEE-2021}. Therefore, many morphing attack detection algorithms (MAD) use image quality as a factor in detecting morphed images \cite{Scherhag-PRNU-TBIOM-2019, Fu2022quality}. Furthermore, most methods require a very tight crop around the input faces and produce, correspondingly, a tightly cropped morph image around the face, which might not resemble the real-world scenarios where the requirements demand a larger cut to be taken or the full head to be included and the shoulders be square on to the camera as per ICAO requirements \cite{ICAO-PortraitQuality-TR-2018}.

To address these shortcomings, we propose StableMorph. StableMorph is designed to produce high-quality morphed images that look crisp and genuine to human observers and quality assessment algorithms alike. StableMorph does not require tightly cropped face images, and more importantly, it is capable of producing wider cuts, i.e., images where the subject's head, shoulders, clothes, and hair are present and without shadow artifacts or pixels warping and manipulation that make them easy to detect by human observers or forensic tools. It also allows for controlling different aspects of the morphed image, such as the color of the eyes, the background, the lighting conditions, adding accessories, and more. These powerful capabilities allow one to tailor the morphed image to the exact specifications of the indented use, producing more challenging and harder-to-detect morph attacks. In summary, our main contributions are:

\begin{itemize}
    \item We introduce StableMorph, a novel face morph generation method that produces morphed images with far superior quality to existing methods and even genuine bona fide input images.
    \item We release high-quality morph datasets for the FRLL \cite{DeBruine2021} dataset, which opens the door for training and evaluating morph detection methods on high-quality morph images.
    \item We present an extensive evaluation study that covers not only the canonical morph attack potential (MAP) metric but also face image quality assessment (FIQA) and general image quality assessment (IQA) for our method and the baseline methods.
\end{itemize}

\section{Related Work}
\label{sec:related-work}

\begin{figure*}[t]
\centering
\begingroup
\setlength{\tabcolsep}{1.3pt} % Default value: 6pt
\begin{tabular}{ccccccc}
    
    Subject 1 &
    FaceMorpher &
    LMA-UBO &
    \textbf{StableMorph} & 
    MIPGAN-I &
    % MIPGAN-II & 
    MorDiff & 
    Subject 2 \\

    \includegraphics[width=.14\linewidth]{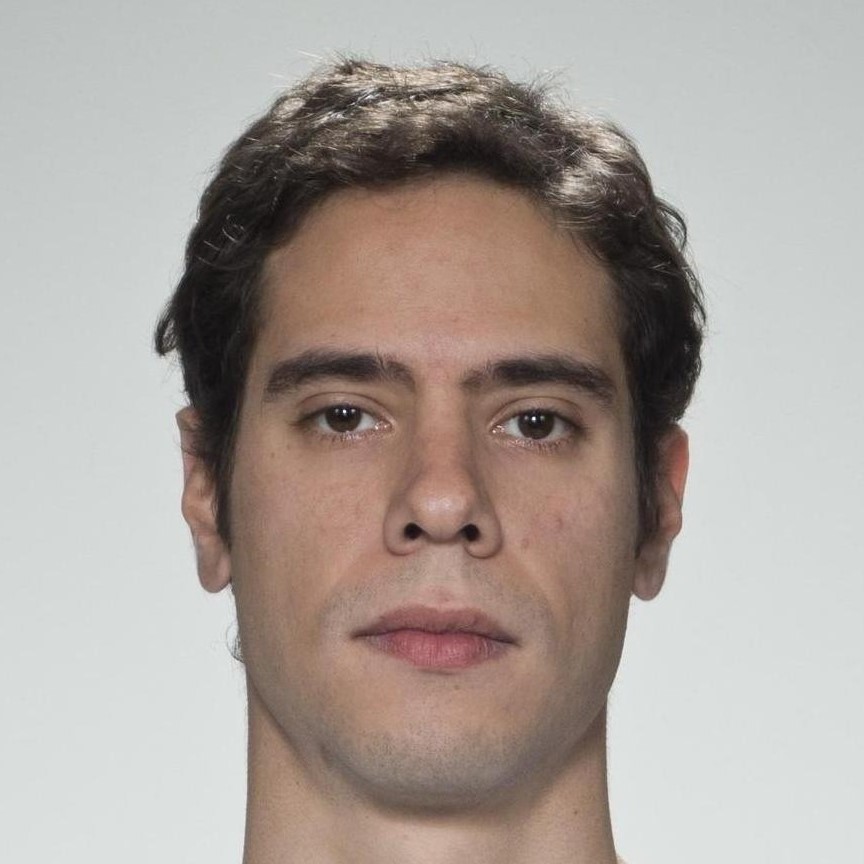} &
    \includegraphics[width=.14\linewidth]{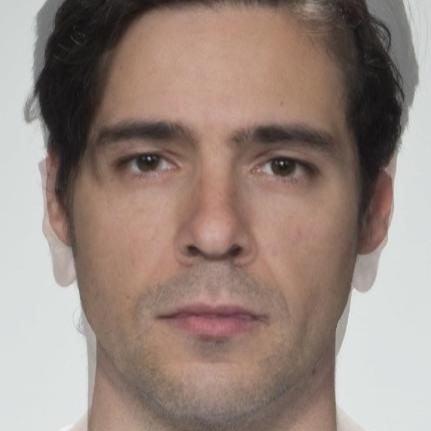} &
    \includegraphics[width=.14\linewidth]{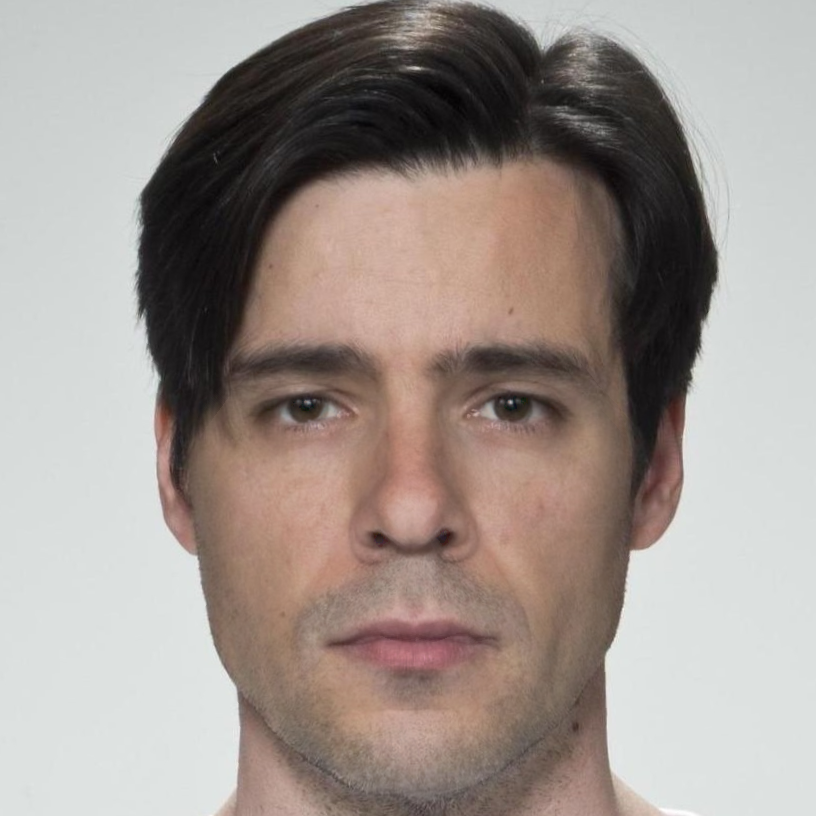} &
    \includegraphics[width=.14\linewidth]{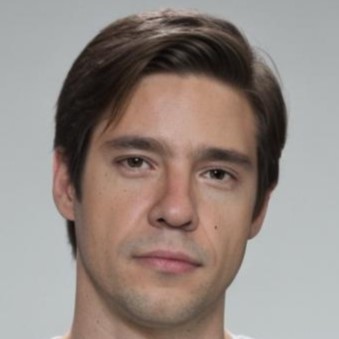} &
    \includegraphics[width=.14\linewidth]{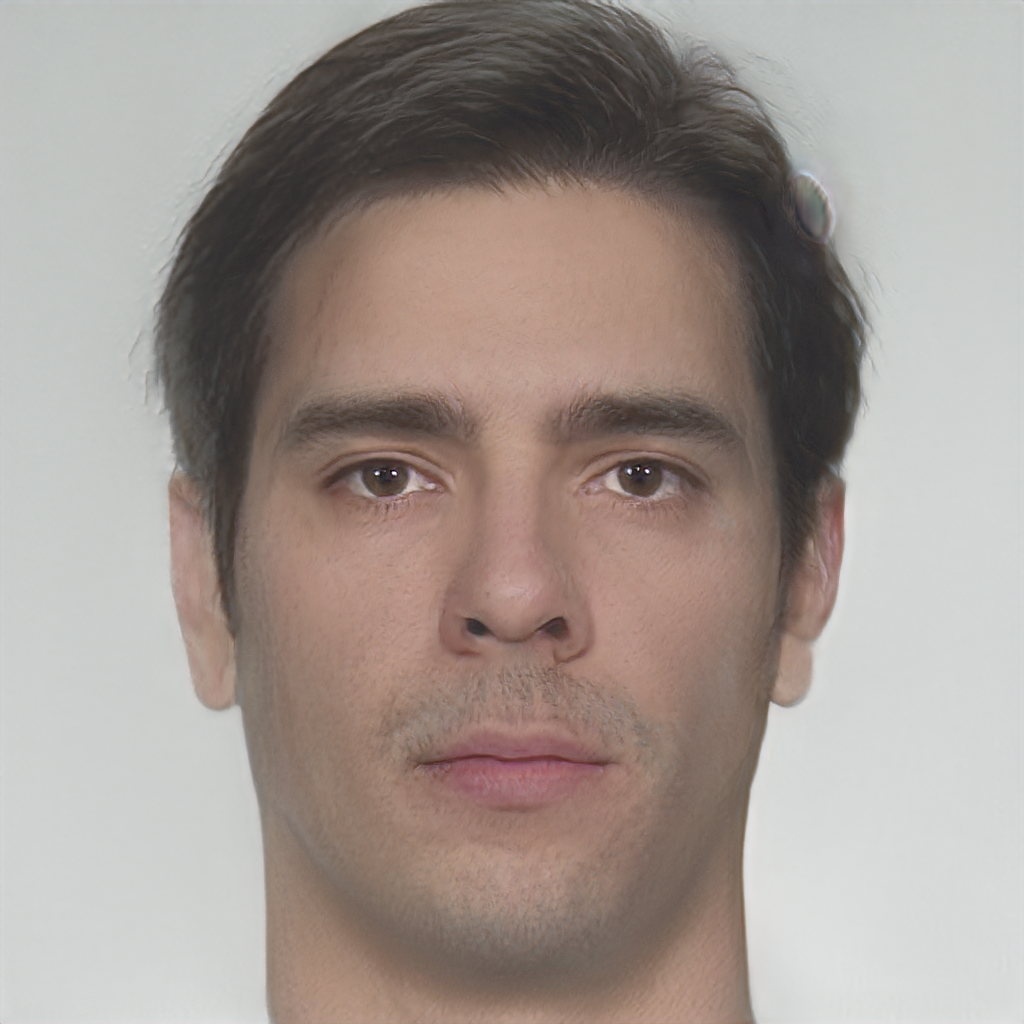} &
    \includegraphics[width=.14\linewidth]{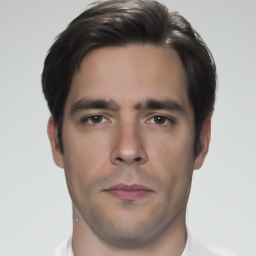} &
    \includegraphics[width=.14\linewidth]{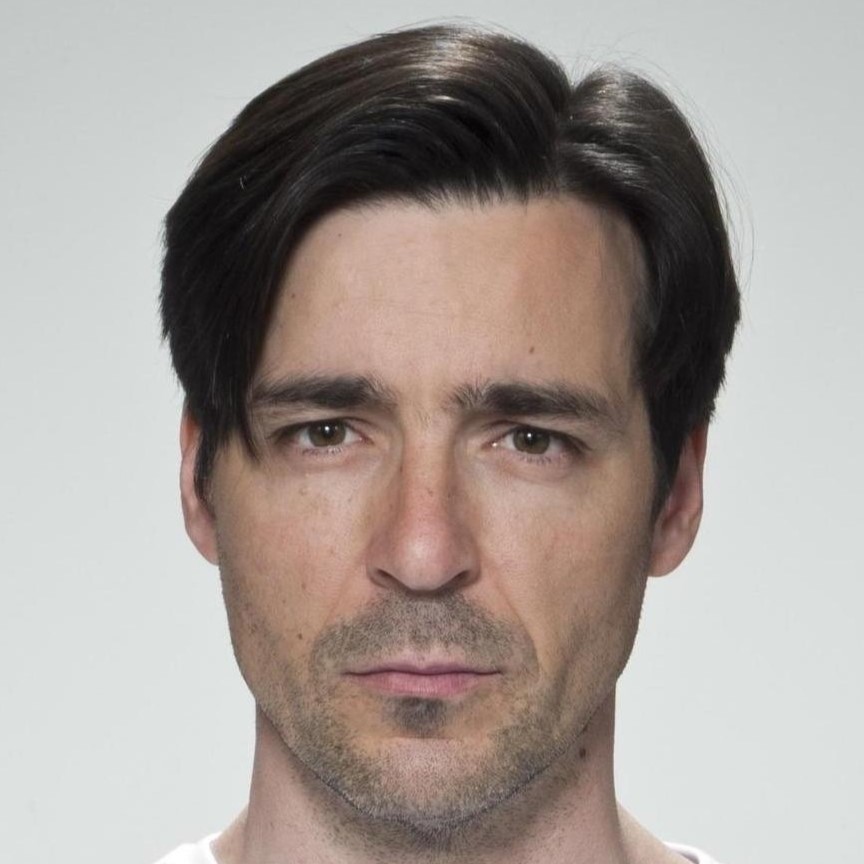} \\

    \includegraphics[width=.14\linewidth]{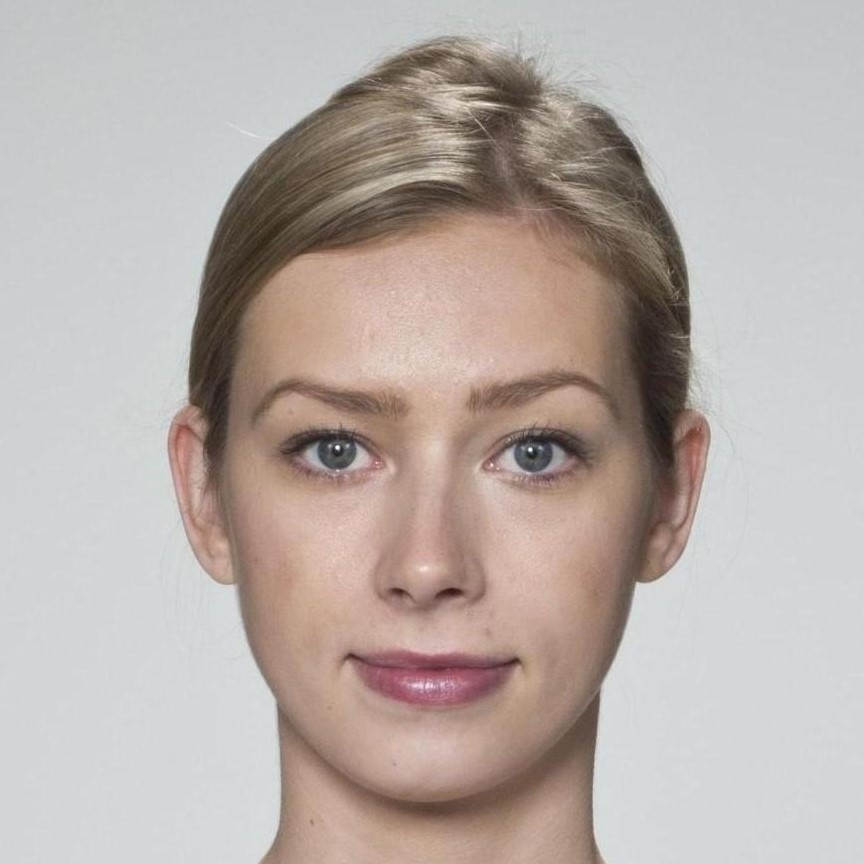} &
    \includegraphics[width=.14\linewidth]{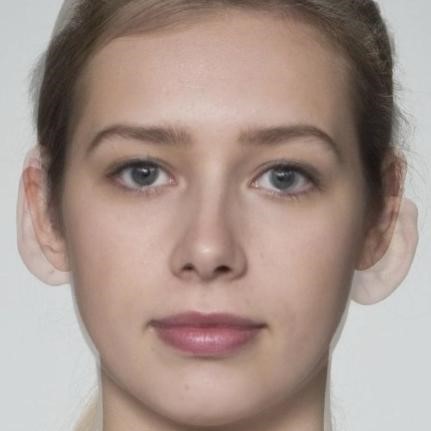} &
    \includegraphics[width=.14\linewidth]{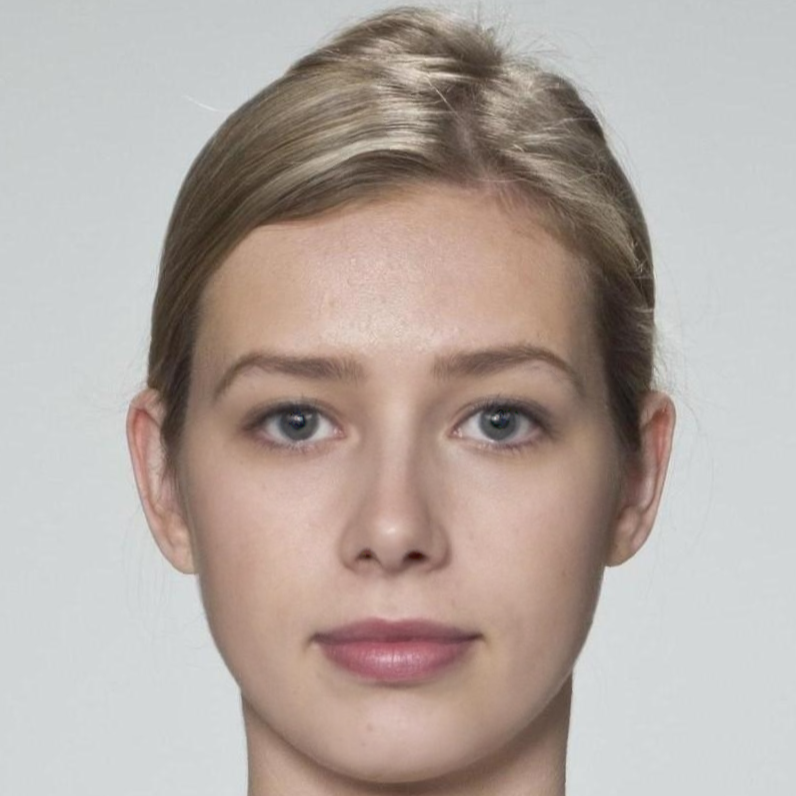} &
    \includegraphics[width=.14\linewidth]{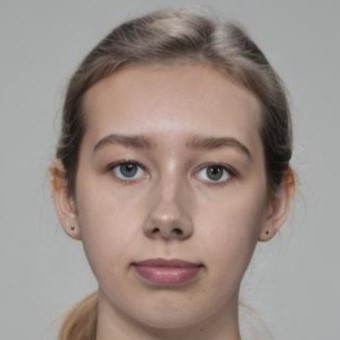} &
    \includegraphics[width=.14\linewidth]{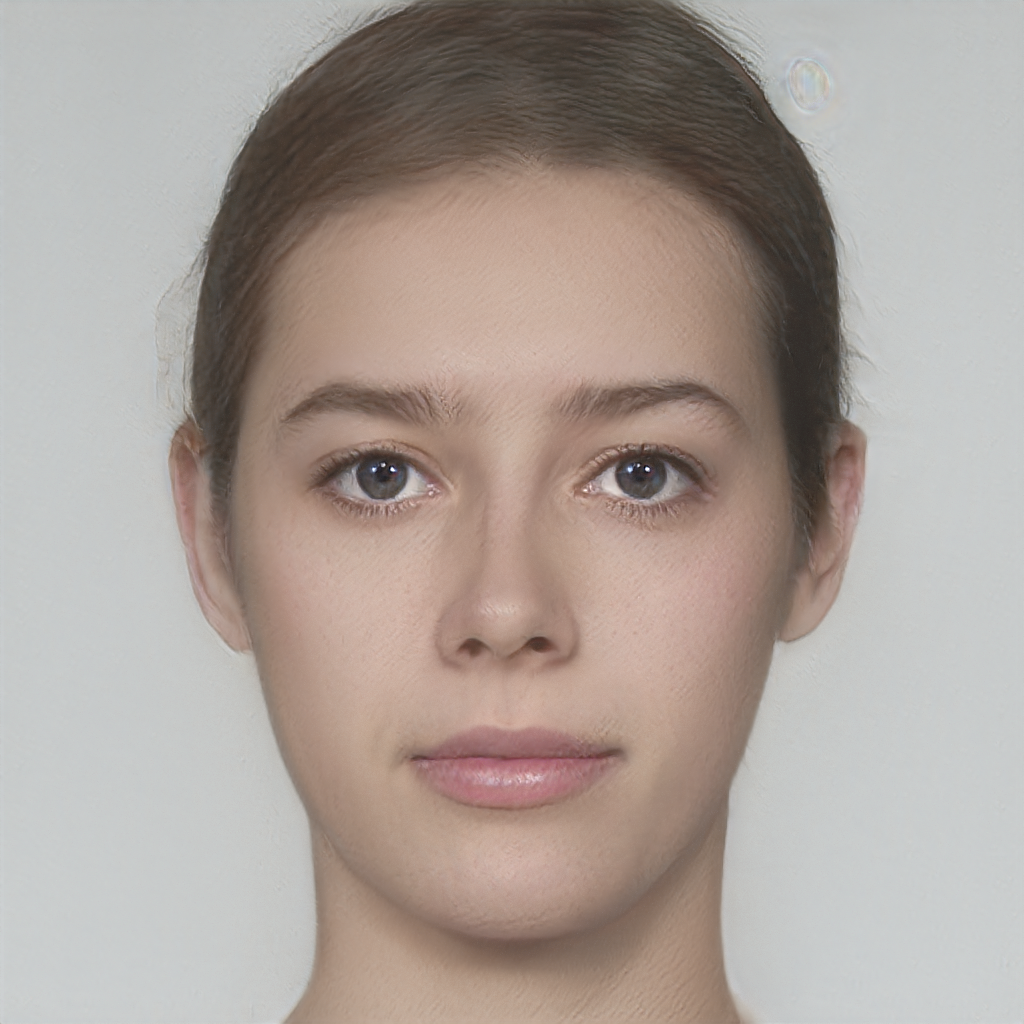} &
    \includegraphics[width=.14\linewidth]{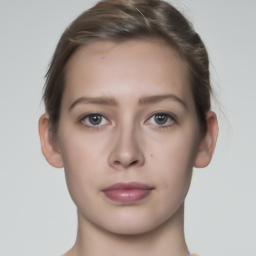} &
    \includegraphics[width=.14\linewidth]{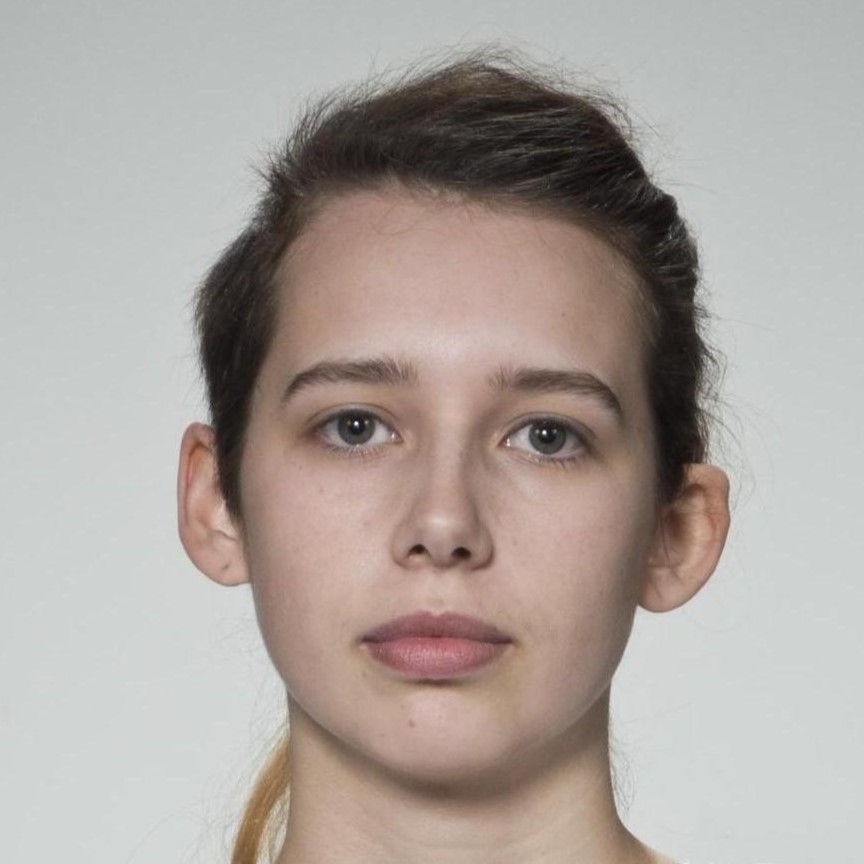} \\

    \includegraphics[width=.14\textwidth]{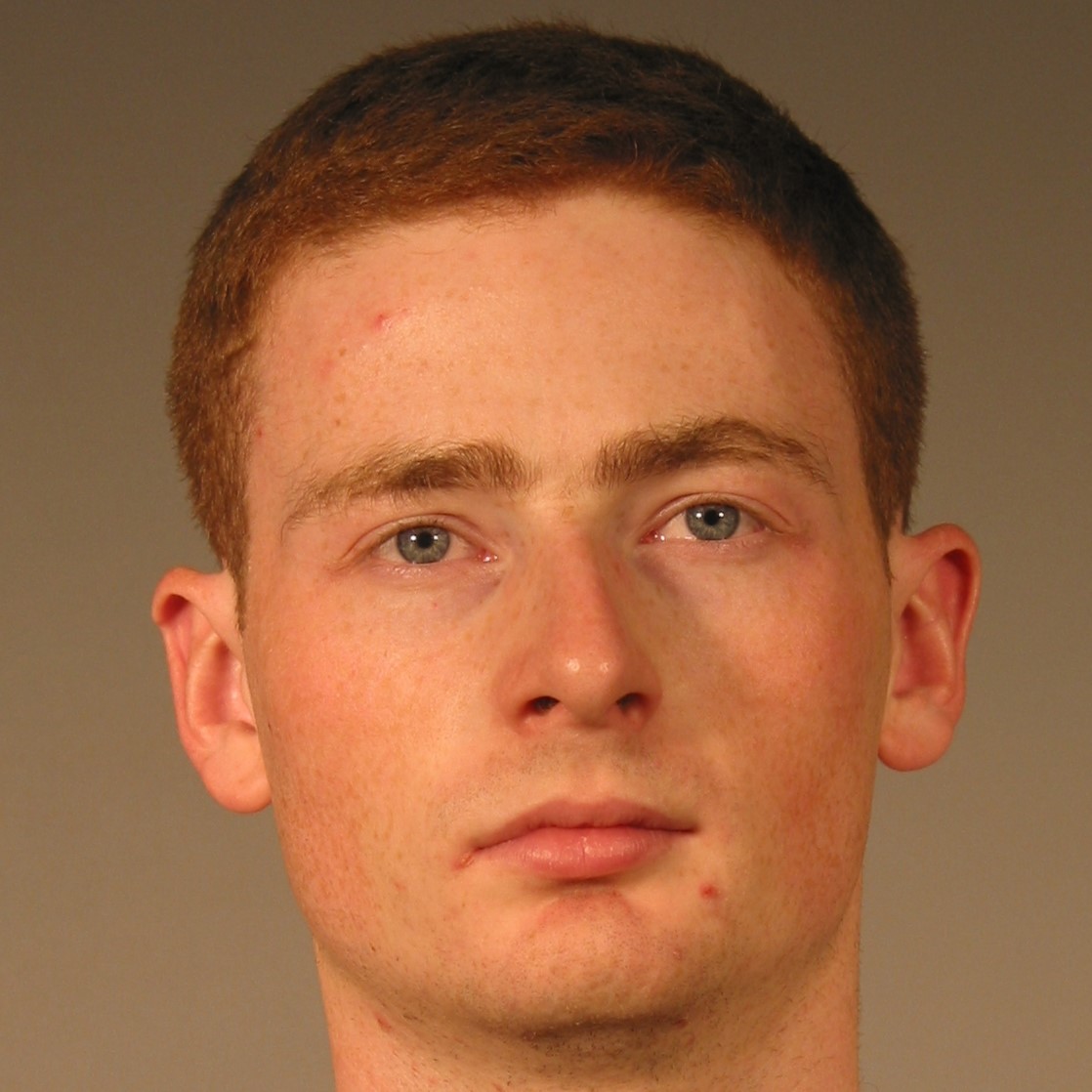} &
    \includegraphics[width=.14\textwidth]{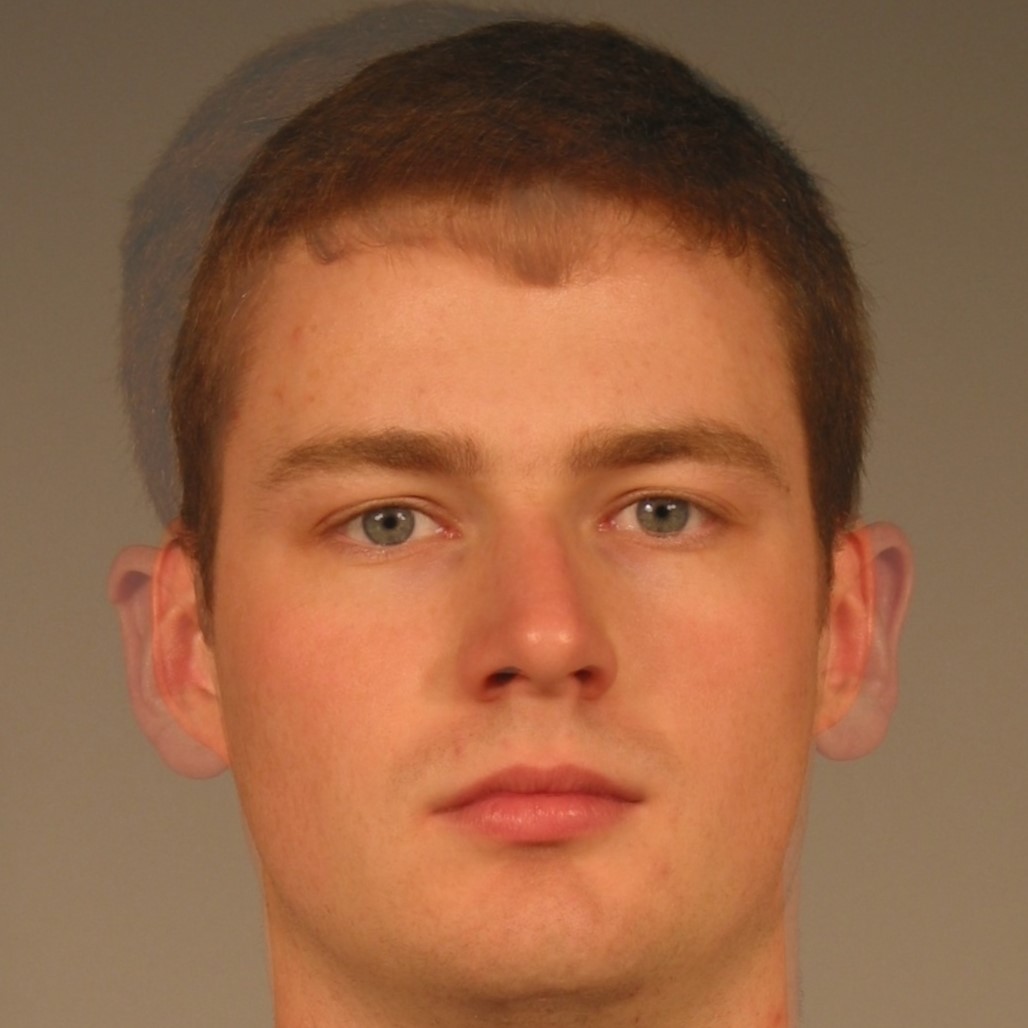} &
    \includegraphics[width=.14\textwidth]{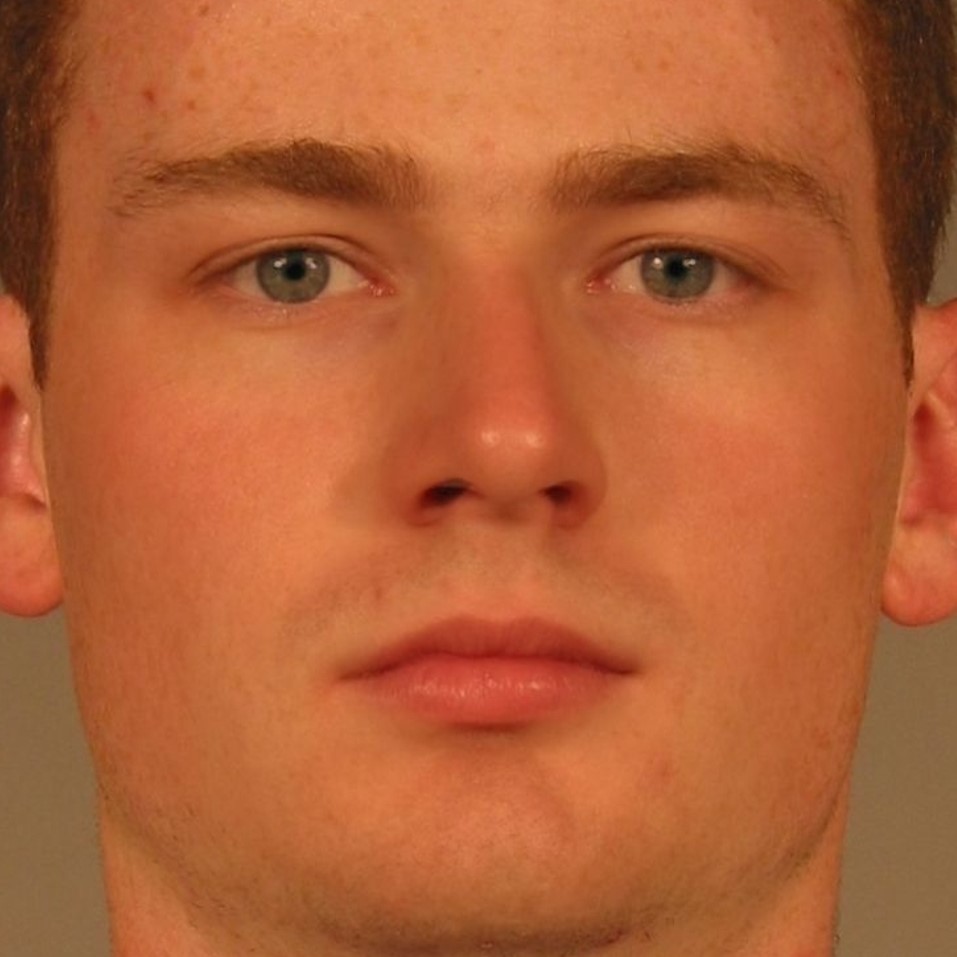} &
    \includegraphics[width=.14\textwidth]{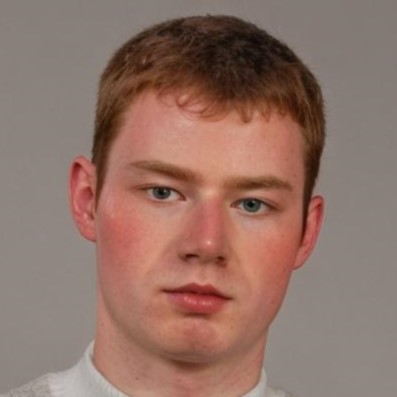} &
    \includegraphics[width=.14\textwidth]{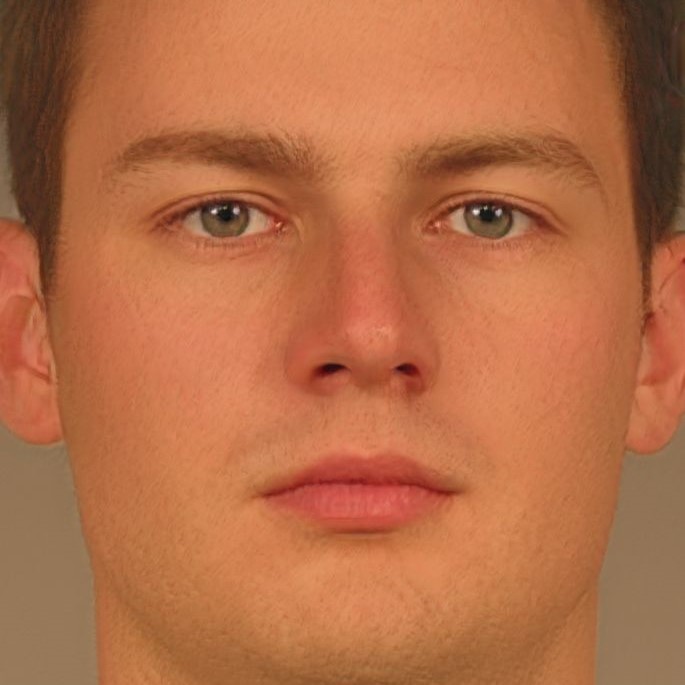} &
    \includegraphics[width=.14\textwidth]{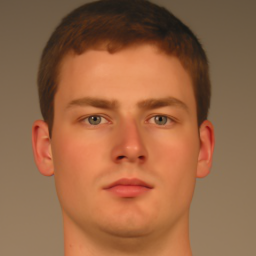} &
    \includegraphics[width=.14\textwidth]{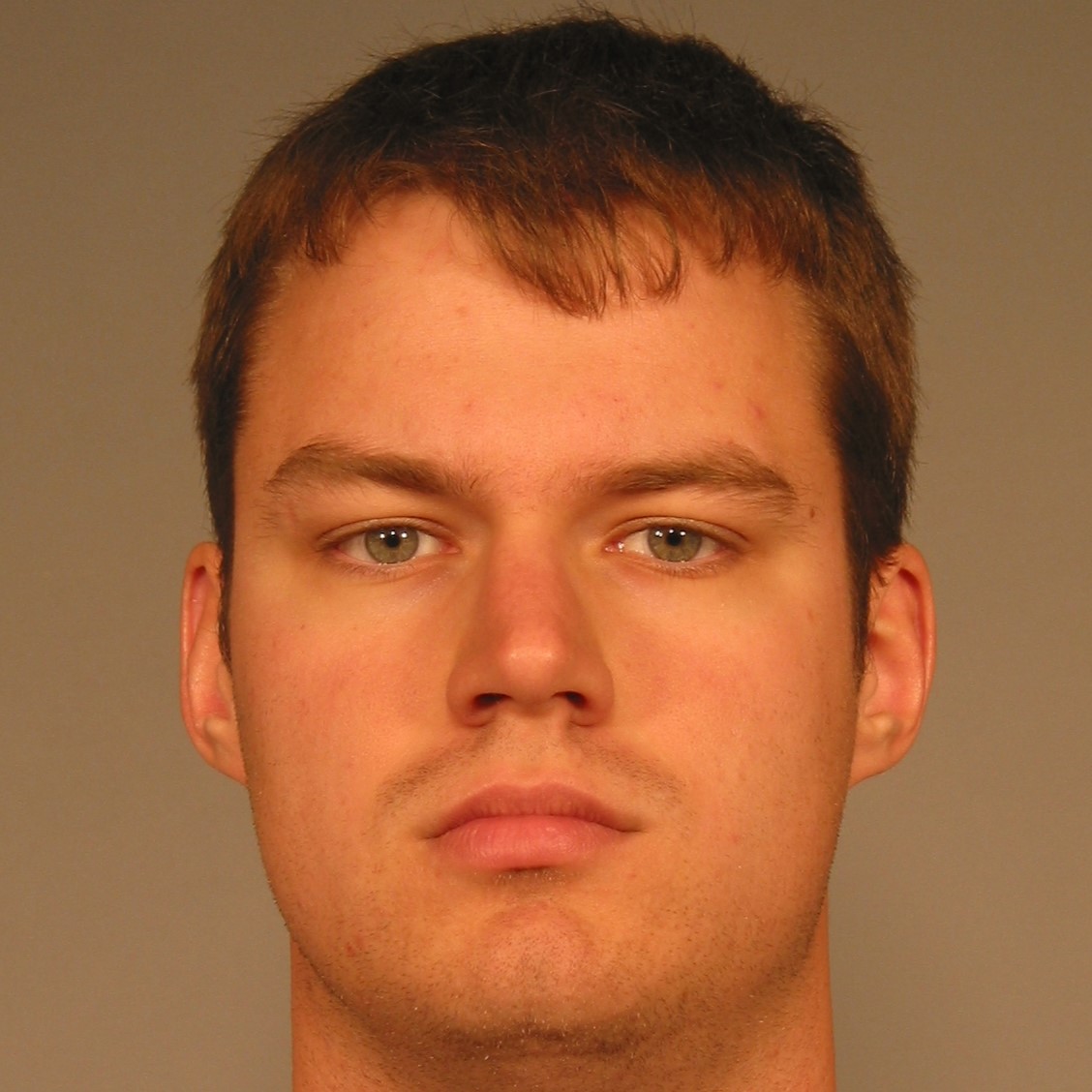} \\

    \includegraphics[width=.14\textwidth]{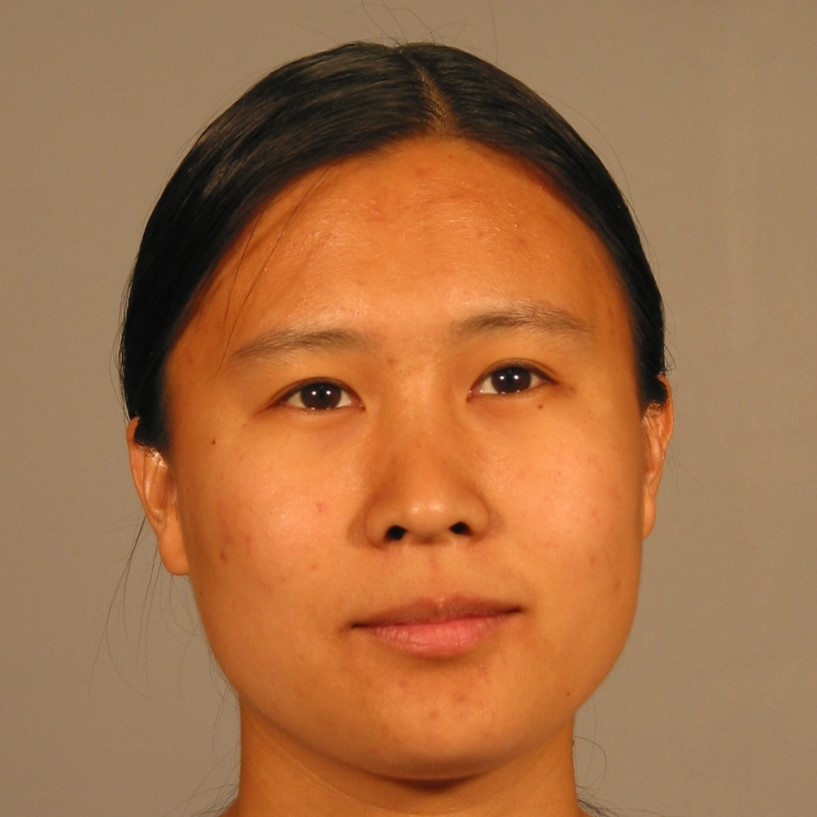} &
    \includegraphics[width=.14\textwidth]{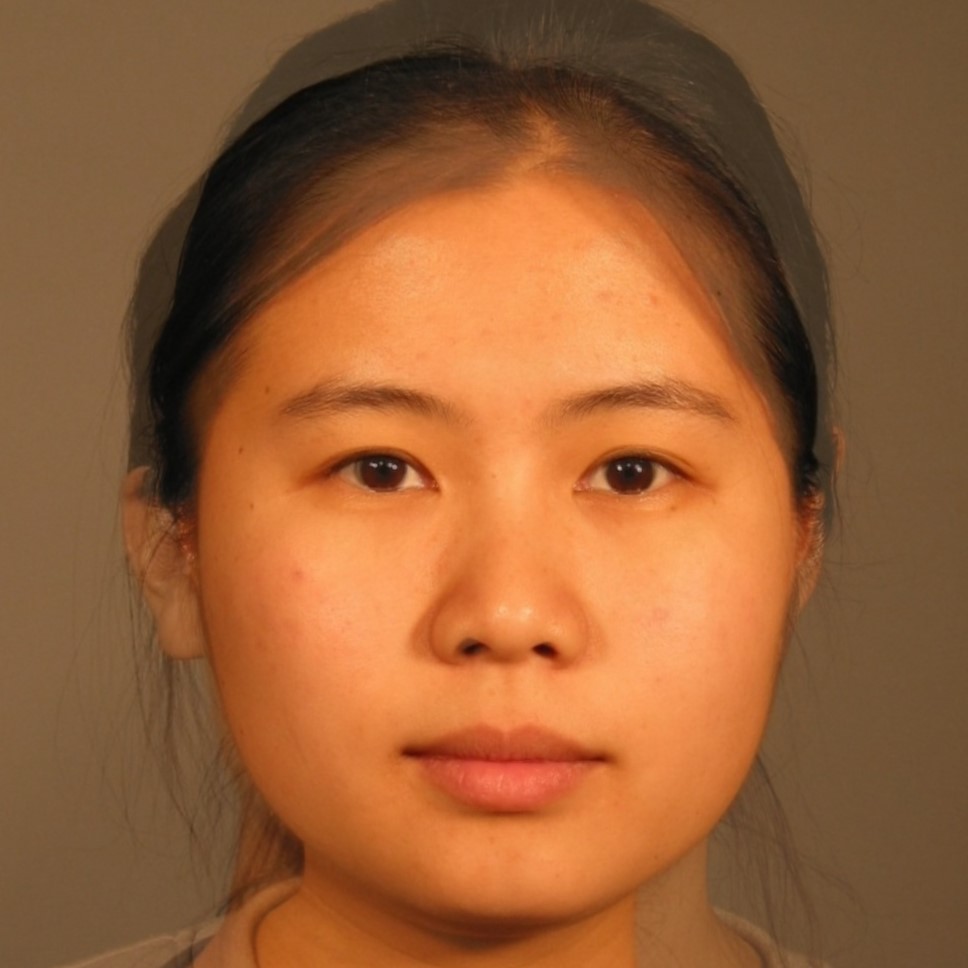} &
    \includegraphics[width=.14\textwidth]{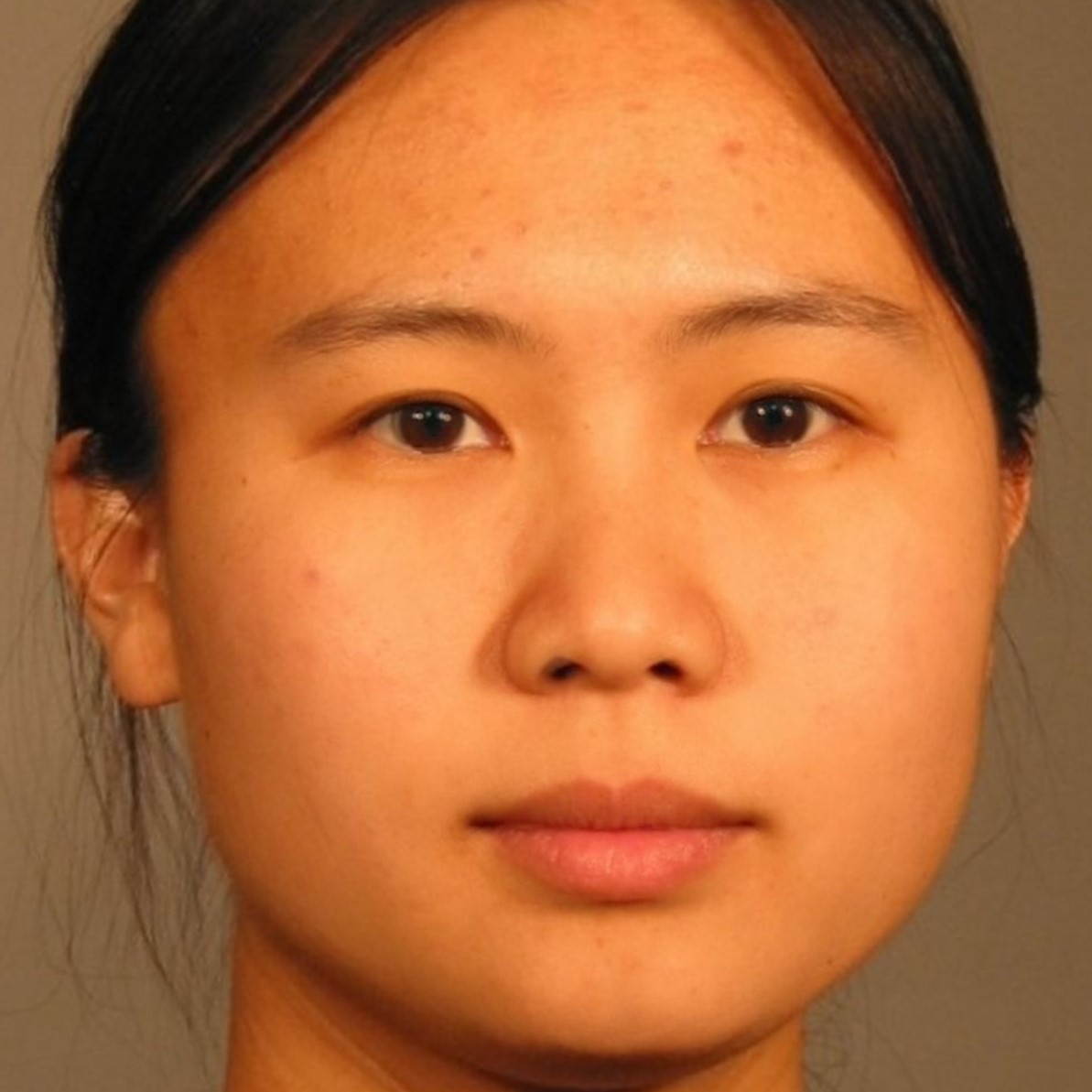} &
    \includegraphics[width=.14\textwidth]{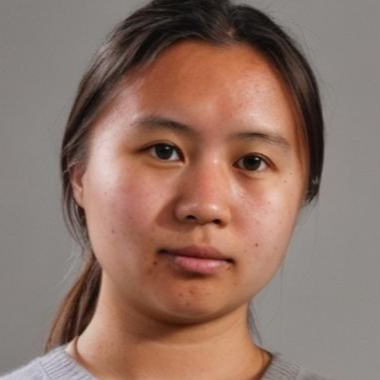} &
    \includegraphics[width=.14\textwidth]{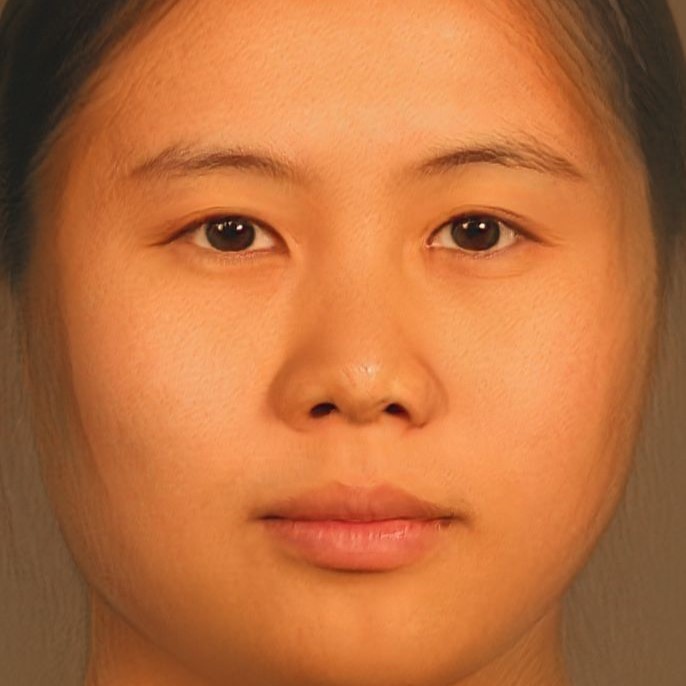} &
    \includegraphics[width=.14\textwidth]{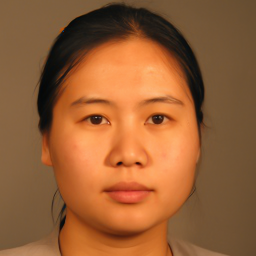} &
    \includegraphics[width=.14\textwidth]{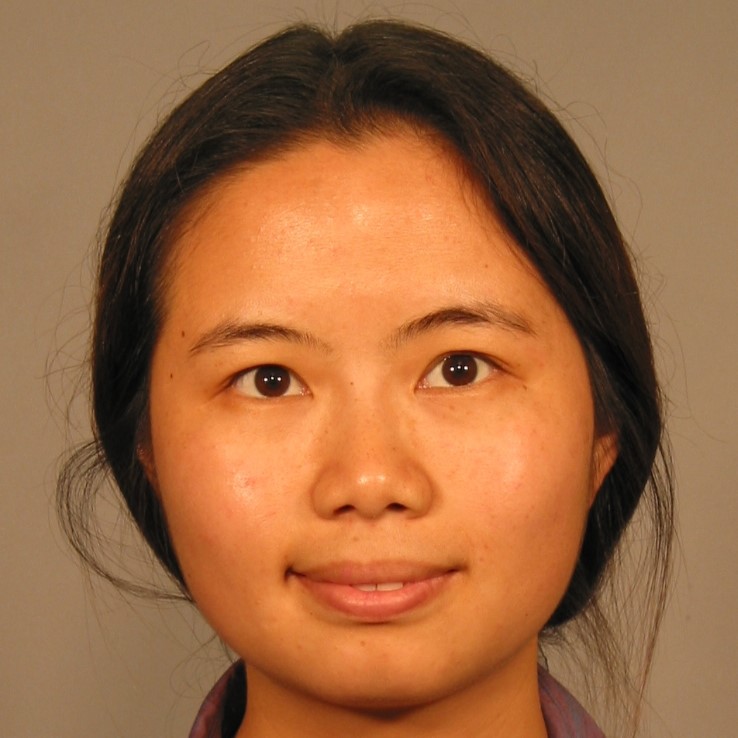}

\end{tabular}
\endgroup
\caption{Comparison of morphed images from the different methods. Images from FRLL are in the top two rows, and from FRGC in the bottom two ones. A nice feature of StableMorph is that it is not influenced by the accent color in the original images.}
\label{fig:qualitative-frll}
\end{figure*}

\subsection{Face Morph Generation}

In addition to the GAN-based morph generation methods, several methods that use diffusion models have recently been proposed. MorDiff uses diffusion autoencoders to create face morphing attacks \cite{Damer2023MorDiff}. LADIMO presents a representation-level face morphing approach that performs morphing on the face embeddings of the contributing subjects using a trained Latent Diffusion Model (LDM) \cite{grimmer2024ladimo}. DiffMorpher, not specifically aimed at face images, proposes an approach that enables smooth and natural image interpolation by harnessing the knowledge of a pre-trained diffusion model \cite{Zhang_2024_CVPR}. A recent approach uses neural networks to parameterize image warpings, embeds the warpings into the latent space of a diffusion autoencoder, then interpolates the resulting embedding and projects it back to the image space \cite{schardong2024neural}.

\subsection{Morph Attack Detection}

Morph attack detection (MAD) algorithms are typically also split into two main categories. Single image-based MAD (S-MAD) and differential image-based MAD (D-MAD). S-MAD algorithms operate on a single image only to decide whether the suspected image is morphed or not. D-MAD algorithms have access to a reference image taken in a trusted environment and guaranteed to be bona fide \cite{Venkatesh-FaceMorphingAttackGenerationAndDetection-TTS-2021}. MAD algorithms use different techniques, such as analyzing the texture features of the image \cite{Raghavendra-DetectingMorphedFace-BTAS-2016, Scherhag-MorphingDetection-ICBEA-2018}, analyzing various quality aspects of the image such as image degradation artifacts, corners and edge distortions \cite{Scherhag-PRNU-TBIOM-2019}, and feature analysis using deep learning models \cite{Venkatesh-FaceMorphingAttackGenerationAndDetection-TTS-2021}.

\subsection{IQA and FIQA}

We cover the concepts of general image quality assessment (IQA) and face image quality assessment (FIQA) because we use some of these methods in our evaluation procedure. Besides, the aspect of face image quality is important to highlight because the FIQA process is becoming an integral part of online ID document issuance applications. Hence, a morphed image can be rejected simply due to its poor quality, even if it was not flagged as a morphed image. For example, by being too grainy, blurry, or having bad illumination.

Image quality assessment (IQA) refers to the process of evaluating the visual or perceptual quality of an image, typically through objective algorithms or subjective human judgments \cite{mittal2012no}. The objective quality methods can be divided into full-reference (FR-IQA) and no-reference (NR-IQA) methods, depending on whether the image is being evaluated relative to a reference image \cite{lao2022attentions, cheon2021perceptual}. No-Reference Image Quality Assessment (NR-IQA) is useful in assessing the perceptual quality of images in accordance with human subjective perception \cite{ying2020patches, yang2022maniqa}. 

Face image quality assessment (FIQA) refers to the process of estimating the utility of a face image for face recognition \cite{Schlett-FIQA-LiteratureSurvey-CSUR-2021}. High-quality images are expected to yield better FR performance \cite{Boutros-CR-FIQA-CVPR-2023}. The ISO/IEC 29794-5 international standard on face image quality defines several \textit{measures} for determining face image quality \cite{ISO-IEC-29794-5-DIS-FaceQuality-240129}. These \textit{measures} are split into two kinds. (1) A \textit{unified quality score (UQS)} that assesses overall image quality without being explicitly tied to a specific aspect. (2) \textit{Quality components} which assess specific aspects of the face image, such as sharpness, background uniformity, illumination uniformity, and under/over exposure \cite{merkle2022state, ISO-IEC-29794-5-DIS-FaceQuality-240129}.

\subsection{Diffusion Models}

Diffusion models (DM) \cite{sohldickstein2015deep} are a class of deep generative models that have recently gained popularity for many tasks in the computer vision domain \cite{Croitoru2023DMV, Yang2023DMA}. They involve a forward process where data is progressively transformed into noise and a reverse process where noise is progressively turned back into data, achieving state-of-the-art synthesis results on image data and beyond \cite{dhariwal2021diffusion, Cao2024GDM}. The initially introduced diffusion models operate directly in the image space, which means that training or fine-tuning them requires extensive time and compute resources \cite{ho2020denoising}. This gave rise to latent diffusion models (LDM), where the diffusion process is applied instead in the latent space of pretrained autoencoders \cite{rombach2022highresolution}. LDMs utilize a UNet architecture as the denoising autoencoder. The UNet's encoder-decoder structure, with skip connections, allows it to effectively capture both local and global features, enhancing its ability to progressively refine and recover the original data from the noisy latent space \cite{rombach2022highresolution}. They have been shown to achieve state-of-the-art results for image inpainting, conditional and unconditional image synthesis, and super-resolution while significantly reducing computational requirements compared to pixel-based DMs \cite{esser2024scaling}.
\section{Proposed Method}
\label{sec:approach}

\begin{figure*}[t]
    \centering
    \includegraphics[width=\linewidth]{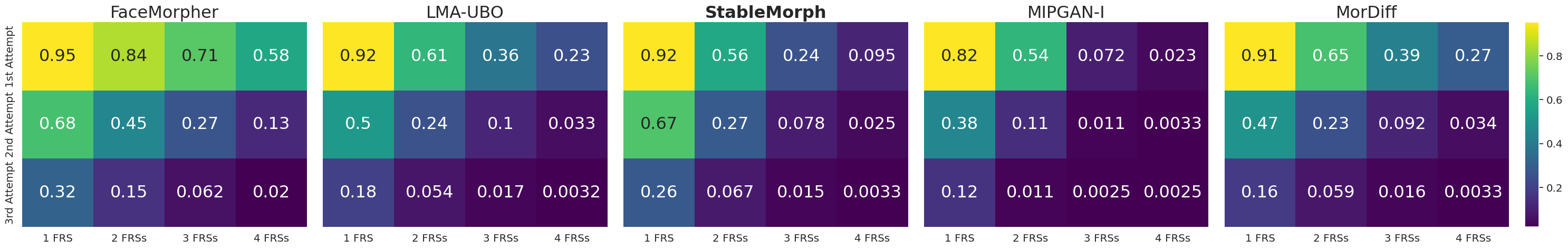}
    \caption{Morph Attack Potential results on FRLL assessed with three verification attempts and four FRSs with a fixed FMR at 0.1\%.}
    \label{fig:map-frll}
\end{figure*}

\begin{figure*}[t]
     \centering
    \includegraphics[width=\linewidth]{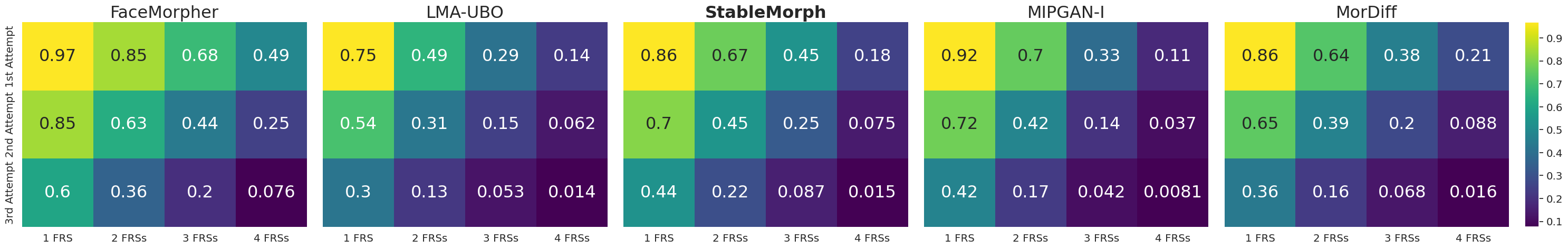}
    \caption{Morph Attack Potential results on FRGC assessed with three verification attempts and four FRSs with a fixed FMR at 0.1\%.}
    \label{fig:map-frgc}
\end{figure*}

The face morphing process can include images of more than two subjects, and this is also possible with our proposed method. However, most works in the literature focus on two subjects only, as it is a more realistic scenario, so we confine the explanation to two subjects to keep the description of the method more concise.

Given $S_1$ and $S_2$ are two distinct subjects. $I_1^n$ is a set of $n$ face images of $S_1$, and $I_2^n$ is a set of $n$ face images of $S_2$, where $n >= 1$. The goal is to generate a high-quality morphed face image $M$ that resembles both $S1$ and $S2$ and can be used to deceive FRSs and human inspection. The approach we propose in StableMorph to achieve this has four stages. First, we fine-tune a pretrained diffusion model on face images of $S_1$ and $S_2$. Second, we extract the identities of $S_1$ and $S_2$ using a FRS. Third, we merge the two fine-tuned models as well as the identities of $S_1$ and $S_2$. Lastly, we use the merged model and the merged identities to generate the morphed image $M$.

\subsection{Fine-tuning}

To make a pretrained diffusion model (DM) learn details about a given subject so it can be used to generate images of the subject, we need to fine-tune it. However, StableMorph is based on \textit{Stable Diffusion (SD)}, a state-of-the-art latent text-to-image diffusion model capable of generating high-resolution, photo-realistic images \cite{rombach2022highresolution}. Despite being an LDM that requires four times less training resources compared to traditional DMs, it still has approximately 860 million parameters \cite{rombach2022highresolution}, which is still unrealistic for our scenario and the computing resources we have.

To overcome this problem, we use Low-Rank Adaptation (LoRA) to achieve the needed fine-tuning. LoRA is a technique initially introduced for fine-tuning large language models. It works by freezing the weights of the pre-trained model and training only a smaller number of newly injected weights into the model \cite{hu2022lora}. This makes training larger models, such as SD, much faster, more memory-efficient, and more feasible for our purpose. Given $I^m$ a set of $m$ face images of subject $S$ where $m >=1$ but not necessarily equals $n$, we use the LoRA technique to fine-tune the UNet component in SD on $I^m$ and produce a set of learned weights $W$. A similar approach is used in DiffMorpher to interpolate between two images, though not specifically face images \cite{Zhang_2024_CVPR}, and in ZipLoRA to merge any any user-provided subject with any user-provided style \cite{shah2023ZipLoRA}.

\subsection{Extracting Face Identities}

To make sure the generated morphed image $M$ resembles the two subjects $S_1$ and $S_2$ and, more importantly, internalizes the identity information of $S_1$ and $S_2$, we incorporate these identities into the generation process. To this end, we need to extract the identities from the face images. Given $I^n$ is a set of $n$ images of subject $S$, we use a face recognition model to extract the face embeddings $E^{(n*d)}$ from $I^n$, where $d$ is the size of the feature vector (embedding) the face recognition model produces.

\subsection{Merging LoRAs and Identities}

Before we can generate the morphed image, we need to first merge the two LoRAs (the two sets of weights learned by the fine-tuning process) and then the two identities. More formally, given that $W_1$ is the set of LoRA weights learned from fine-tuning SD on $I_1^m$ and $W_2$ is the set of LoRA weights learned from fine-tuning SD on $I_2^m$, we merge $W_1$ and $W_2$ to produce $W_m$ as shown in Equation \ref{eq:merging-weights}.

\begin{equation} 
W_m = 0.5 * W_1 + 0.5 * W_2
\label{eq:merging-weights}
\end{equation}

The merged weights $W_m$ are then loaded into the UNet part of the original SD model, creating a modified version $SD_m$ ready for generating the morphed image $M$. To merge the identities of $S_1$ and $S_2$, we merge the two embedding matrices $E_1^{(n*d)}$ and $E_2^{(n*d)}$ using Spherical Interpolation (SLERP) with $\lambda = 0.5$, as shown in Equation \ref{eq:slerp}, and obtain a new matrix $E_m^{(n*d)}$ containing identity information from $S_1$ and $S_2$.

\begin{equation} 
\begin{gathered}
\label{eq:slerp}
\text{SLERP}(E_1, E_2, \lambda) = \frac{\sin [(1 - \lambda) \Omega]}{\sin \Omega} \cdot E_1 + \frac{\sin [\lambda \Omega]}{\sin \Omega} \cdot E_2 \\
\Omega = \frac{\arccos [E_1 \cdot E_2]}{\lVert E_1 \lVert \cdot \lVert E_2 \lVert}
\end{gathered}
\end{equation}

\subsection{Morphed Image Generation}

To make sure the modified model $SD_m$ generates a morphed image $M$ aligned with the identities of $S_1$ and $S_2$, we condition the generation process on the merged embedding matrix $E_m^{(n*d)}$. To this end, we use \textit{IP-Adapter}, an effective and lightweight adapter designed to enable a pretrained text-to-image diffusion model to generate images with an image prompt \cite{ye2023ip-adapter}. IP-Adapter consists of two parts: an image encoder to extract image features from the image prompt and adapted modules with decoupled cross-attention to embed image features into the pretrained text-to-image diffusion model \cite{zhang2023adding}. In our case, we use IP-Adapter with no image prompt since we have already embedded image features from both subjects into the model using the learned LoRA weights. Therefore, we use IP-Adapter with $SD_m$ only to plug the identity information $E_m^{(n*d)}$ as the conditional information. Finally, we run the reverse diffusion process to generate one or several images.

\section{Experiments}
\label{sec:experiments}

% \usepackage{multirow}
\begin{table*}[h]
\resizebox{\linewidth}{!}{%
% \begin{center}
\centering
\begin{tabular}{llccccccc}
\toprule
\multicolumn{1}{l}{Method} &
  \multicolumn{1}{l}{Dataset} &
  \multicolumn{1}{c}{Bonafide} &
  \multicolumn{1}{c}{FaceMorpher} &
  \multicolumn{1}{c}{LMA-UBO} &
  \multicolumn{1}{c}{MIPGAN-I} &
  \multicolumn{1}{c}{MIPGAN-II} &
  \multicolumn{1}{c}{MorDIFF} &
  \multicolumn{1}{c}{\textbf{StableMorph}} \\
\toprule
\multirow{2}{*}{Sharpness $\uparrow$} 
    & FRLL 
    & 95.70$\pm$10.40 
    & 19.71$\pm$15.77 & 39.76$\pm$19.60 & 58.40$\pm$29.34 & 30.06$\pm$22.57 & 75.32$\pm$21.83 & \textbf{96.50$\pm$8.85} \\
   & FRGC 
   & 68.23$\pm$41.81 
   & 44.20$\pm$16.11 & 43.56$\pm$19.83 & 29.65$\pm$15.68 & 34.64$\pm$17.77 & 44.47$\pm$21.28 & \textbf{75.51$\pm$26.59} \\
\midrule
\multirow{2}{*}{UQS $\uparrow$}       
    & FRLL 
    & 56.56$\pm$21.00 
    & 61.02$\pm$11.74 & 68.18$\pm$10.75 & 43.57$\pm$12.19 & 44.37$\pm$12.37 & 71.28$\pm$11.20 & \textbf{85.40$\pm$8.30} \\
   & FRGC 
   & 61.90$\pm$17.56 
   & 58.45$\pm$12.98 & 61.99$\pm$12.73 & 48.66$\pm$12.43 & 47.32$\pm$12.06 & 68.77$\pm$19.20 & \textbf{85.42$\pm$9.52} \\
\midrule

\multirow{2}{*}{SDD-FIQA $\uparrow$}
    & FRLL 
    & 73.57$\pm$3.86 
    & 76.81$\pm$2.45 & 77.63$\pm$2.75 & 74.66$\pm$2.64 & 74.45$\pm$2.56 & \textbf{79.28$\pm$2.52} & 78.04$\pm$3.03 \\
   & FRGC 
   & 74.23$\pm$4.81 
   & 76.40$\pm$2.63 & 71.91$\pm$3.83 & 69.67$\pm$4.80 & 69.56$\pm$4.23 & 77.37$\pm$4.57 & \textbf{77.80$\pm$2.84} \\
\midrule

\multirow{2}{*}{PaQ2PiQ $\uparrow$}   
    & FRLL 
    & 0.41$\pm$0.02   
    & 0.45$\pm$0.01   & 0.38$\pm$0.01   & 0.36$\pm$0.02   & 0.38$\pm$0.01  & \textbf{0.57$\pm$0.02} & 0.54$\pm$0.02   \\
   & FRGC 
   & 0.34$\pm$0.06   
   & 0.36$\pm$0.02   & 0.39$\pm$0.02   & 0.47$\pm$0.02   & 0.48$\pm$0.02   & 0.52$\pm$0.06 & \textbf{0.55$\pm$0.03}   \\
\midrule
\multirow{2}{*}{MANIQA $\uparrow$}    
    & FRLL 
    & 70.47$\pm$2.54  
    & 74.00$\pm$0.98  & 69.81$\pm$2.38  & 69.17$\pm$1.68  & 69.57$\pm$1.73  & \textbf{76.33$\pm$0.88} & 75.33$\pm$1.05  \\
   & FRGC 
   & 68.73$\pm$6.93  
   & 68.97$\pm$1.98  & 65.92$\pm$3.47  & 74.01$\pm$0.95  & 74.10$\pm$1.00  & \textbf{76.64$\pm$1.86} & 76.10$\pm$0.99  \\
\midrule
\multirow{2}{*}{FID $\downarrow$}       
    & FRLL 
    & -               
    & 177.38          & \textbf{105.10}          & 175.25          & 180.23          & 147.87 & 118.33 \\
   & FRGC 
   & -               
   & 211.75          & 229.83          & 297.65          & 299.91   & 225.70        & \textbf{153.09} \\  
\bottomrule
\end{tabular}
}
\caption{Summary of the quantitative results with mean$\pm$std for all methods. MorDiff and StableMorph clearly have better image quality compared to other methods, with StableMorph having superior results compared to MorDiff on sharpness, UQS (MagFace), and FID.}
\label{tab:quantitave-table-results}
\end{table*}

\begin{figure*}[h]
    \centering
    \includegraphics[width=\linewidth]{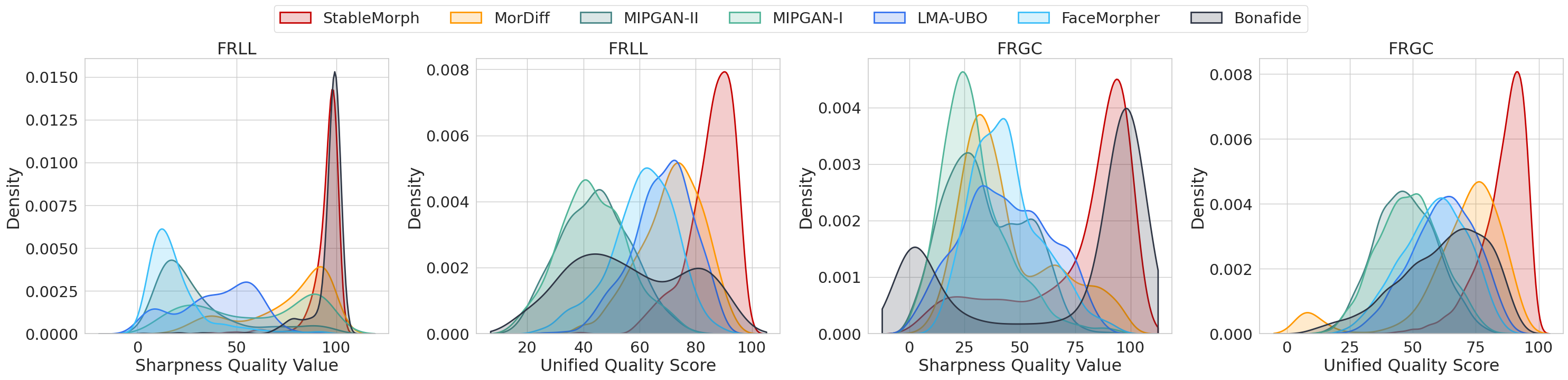}
    \caption{Quality value distributions of the two FIQA measures: sharpness and UQS (MagFace) on FRLL and FRGC. StableMorph clearly has the best sharpness and face image quality of all other methods, including the SoTA MorDiff method.}
    \label{fig:quantitative-fiqa}
\end{figure*}

\begin{figure*}[h]
    \centering
    \includegraphics[width=\linewidth]{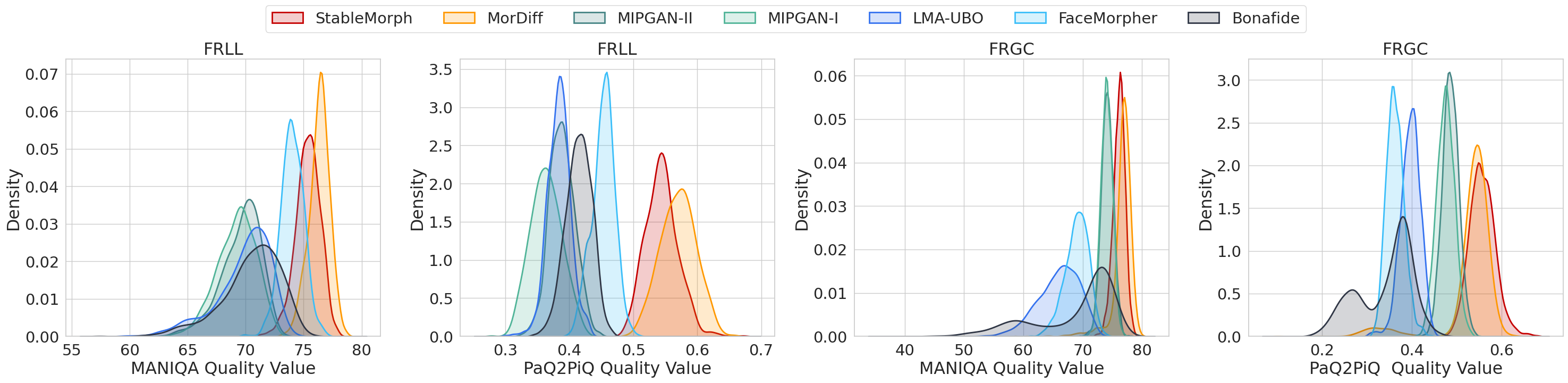}
    \caption{Quality value distributions of the two NR-IQA methods: MANIQA and PaQ2PiQ on FRLL and FRGC with StableMorph and MorDiff having the best overall image quality.}
    \label{fig:quantitative-iqa}
\end{figure*}

\subsection{Experimental Setup}
\label{subsec:setup}

\subsubsection{Datasets}
\label{subsec:datasets}

We use the FRLL \cite{DeBruine2021} and FRGCv2 \cite{Phillips-OverviewFaceRecognitionGrandChallengeFRGC-CVPR-2005} datasets for the experiments. The pairs are selected taking the subject's gender into consideration and using only frontal images with neutral expression. Face embeddings are extracted using ArcFace \cite{Deng-ArcFace-IEEE-CVPR-2019} and the most similar subjects are then selected based on the cosine similarity between the embeddings. An overview of both datasets is shown in Table \ref{tab:datasets}.

\begin{table}[h]
  \centering
  \begin{tabular}{l c c c c}
    \toprule
    Dataset & \#Subjects & \#Images & \#Pairs \\
    \midrule
    FRLL & 102 & 1020 & 1222 \\
    FRGC & 135 & 28910 & 2500 \\
    \bottomrule
  \end{tabular}
  \caption{Overview of the datasets used in the experiments, with the number of pairs used to generate morphed images.}
\label{tab:datasets}

\end{table}

\subsubsection{Baselines}
\label{subsec:baselines}

To evaluate our method against different morph generation techniques, we use two landmark-based methods, FaceMorpher \footnote{https://github.com/yaopang/FaceMorpher} and LMA-UBO \cite{Ferrara-TheMagicPassport-IJCB-2014}, two deep-learning-based methods, MIPGAN-I and MIPGAN-II \cite{Zhang-MIPGAN-MorphingAttacks-IEEE-2021}, and one diffusion autoencoders based method, MorDiff \cite{Damer2023MorDiff} as baselines.

\subsubsection{Implementation Details}

In this work, we use SD v1.5 \cite{rombach2022high}, which generates images at a resolution of 512x512 and can be run and fine-tuned on our hardware. For fine-tuning, we use only one image per subject, i.e., $m = 1$. This is because we are primarily interested in frontal images where the subject is looking directly at the camera with a neutral expression. We use 10 images to extract the identity of the subject, i.e., $n = 10$. This is because we found this to give the best results, and we had this number of images per subject available in the used datasets. Fine-tuning the SD model on a single subject takes approximately 33 minutes and the generation of a single morphed image takes approximately 60 seconds.

\subsection{Quantitative Evaluation}
\label{subsec:quantitative}

\subsubsection{Evaluation Metrics}

We use three kinds of evaluation metrics. First, to evaluate the attack potential, we use the Morphing Attack Potential (MAP) metric introduced by Ferrara et al. \cite{Ferrara-MAP-IWBF-2022} and standardized in the ISO/IEC CD 20059 \cite{ISO-IEC-20059}. As described in the standard, MAP is \textit{a measure of the capability of a morphing attack to deceive one or more biometric recognition systems using multiple recognition attempts.} We use four state-of-the-art FRSs to produce the MAP results: MagFace \cite{Meng-FRwithFQA-MagFace-CVPR-2021}, AdaFace \cite{kim2022adaface}, ArcFace \cite{Deng-ArcFace-IEEE-CVPR-2019}, and SFace \cite{zhong2021sface}. Second, to evaluate the quality of the produced images as face images, we use two FIQA measures: sharpness and a unified quality score. To this end, we use the recently introduced OFIQ framework \footnote{https://github.com/BSI-OFIQ/OFIQ-Project}, which is the reference implementation for the ISO/IEC 29794-5 standard \cite{ISO-IEC-29794-5-DIS-FaceQuality-240129} and SDD-FIQA \cite{ou2021sdd}. Third, to evaluate the overall perceptual quality of the image, we use two methods: MANIQA \cite{yang2022maniqa} and PaQ-2-PiQ \cite{ying2020patches}. Both are recently introduced no-reference IQA methods that aim to assess the perceptual quality of images in accordance with human subjective perception. We also use the canonical Frechet Inception Distance (FID) \cite{heusel2017gans} metric to measure the discrepancy between real bona fide images and generated morphed images. We specifically use the more stable implementation of FID introduced in \textit{clean-fid} \cite{parmar2022aliased} to compute the metric.

\subsubsection{Evaluation Results}

Figures \ref{fig:map-frll} and \ref{fig:map-frgc} show the MAP results on the FRLL and FRGC datasets, respectively. The way MAP results are interpreted is that the value in a cell $[r,c]$ represents the percentage of morphed images successfully exceeding the match threshold and is considered a match against each of the subjects contributing to the morphed image in at least $r$ verification attempts and by at least $c$ of the evaluated FRSs. For example, the top left cell in each matrix will be the percentage of morphed images that are successfully considered a match to both contributing subjects in at least one verification attempt on at least one of the four FRSs. As it can be seen from the results, StableMorph has comparable attack potential to the other methods, with slightly lower attack potential on FRLL and much better resiliency against an increasing number of FRSs on FRGC.

Figure \ref{fig:quantitative-fiqa} shows the distributions of quality values of the two FIQA measures, sharpness and UQS, for the bona fide and the morphed images generated by the different methods. It is clear from the plots that StableMorph has as sharp images as the bona fide ones and superior overall UQS to images produced by the other morphing methods and even to the bona fide ones. Correspondingly, Figure \ref{fig:quantitative-iqa} shows the score distributions for the two IQA methods. It is also clear that StableMorph images have far better perceptual quality than baseline methods and even original bona fide images.

Table \ref{tab:quantitave-table-results} summarizes the results of the FIQA, IQA, and FID metrics with the mean and standard deviation of the scores on FRLL and FRGC. The results show that StableMorph has better results on all metrics, ranking second with a small margin only on FID for FRLL, otherwise having a clear wide margin advantage.

\subsection{Qualitative Evaluation}
\label{subsec:qualitative}

Figure \ref{fig:qualitative-frll} shows sample comparison images from the FRLL and FRGC datasets. The ghost artifacts in FaceMorpher's images are very clearly visible. MIPGAN I images appear washed out, have visible shadow artifacts, and a slight zoom or inspection of the face or the hair reveals how grainy and unnatural the images are. LMA-UBO takes the face frame (head, hair, and ears) from one of the two images and morphs only the face area; thus, it cannot accommodate the shape of the head, ears, or other hair styles and colors. StableMorph images, on the other hand, look natural, genuine, and crisp. StableMorph can accommodate the shape of the head and ears and is not restricted by the hair style, with no shadow effects or visible artifacts around the hair or the face. Further, despite the fact that the LMA-UBO images appear at first glance to be decent, a slight inspection of the images immediately reveals morphing artifacts.

\begin{figure}[t]
\centering
\begingroup
\setlength{\tabcolsep}{1.3pt} % Default value: 6pt
\begin{tabular}{cc}
    
    LMA-UBO &
    \textbf{StableMorph} \\ 
    % LMA-UBO &
    % \textbf{StableMorph} \\
    
    \includegraphics[width=.48\linewidth]{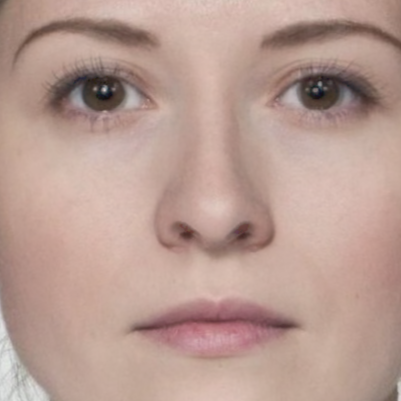} &
    \includegraphics[width=.48\linewidth]{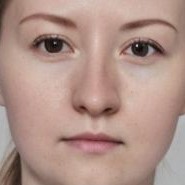} \\

\end{tabular}
\endgroup
\caption{Sample morphed images with UBO-LMA on the left and StableMorph on the right illustrating the morphing artifacts around the iris and the nostrils of the nose in LMA-UBO images vs. the sharp and clean StableMorph images.}
\label{fig:qualitative-artefacts}
\end{figure}

Figure \ref{fig:qualitative-artefacts} shows two images from LMA-UBO and StableMorph. The morphing artifacts with the double iris in the eyes and blurry nostrils of the nose are clearly visible in LMA-UBO. StableMorph's images, on the other hand, are clean, crisp, and sharp even when zoomed in.

\begin{figure*}[ht]
\centering
\begingroup
\setlength{\tabcolsep}{1.3pt} % Default value: 6pt
\begin{tabular}{ccccccc}
    
    Subject 1 &
    (A) &
    (B) &
    Default & 
    (C) &
    (D) & 
    Subject 2 \\

    \includegraphics[width=.14\linewidth]{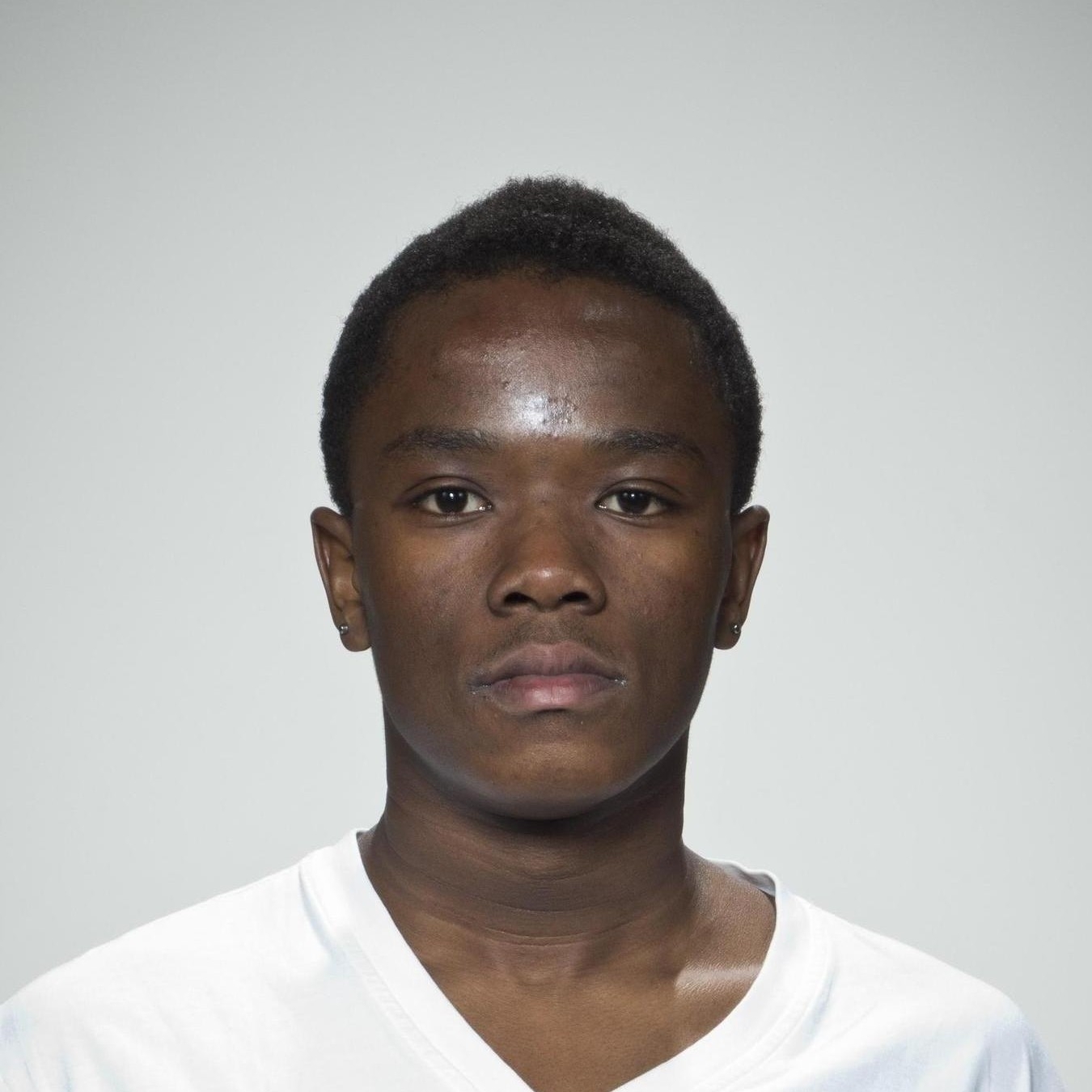} &
    \includegraphics[width=.14\linewidth]{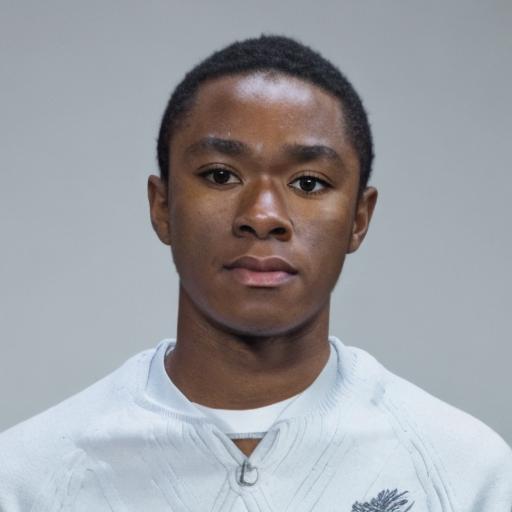} &
    \includegraphics[width=.14\linewidth]{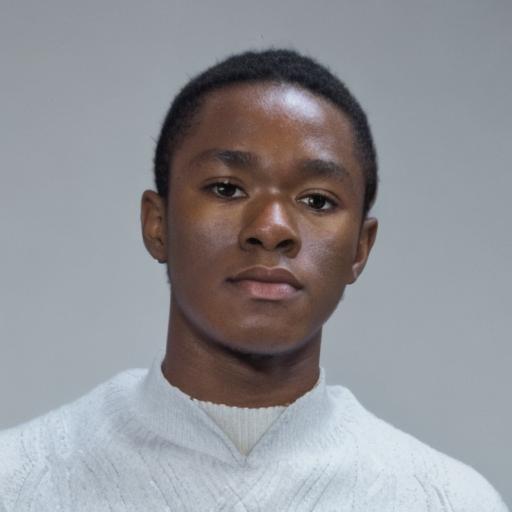} &
    \includegraphics[width=.14\linewidth]{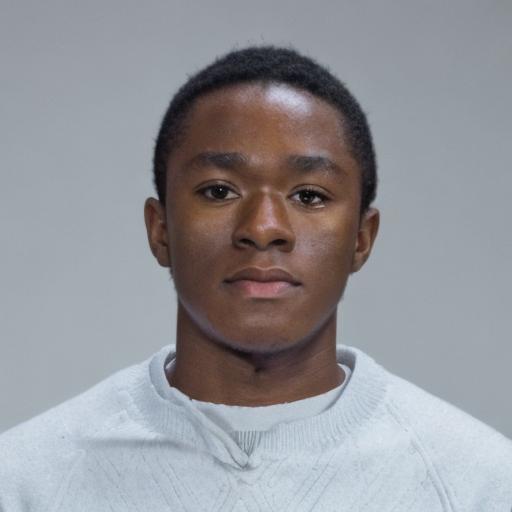} &
    \includegraphics[width=.14\linewidth]{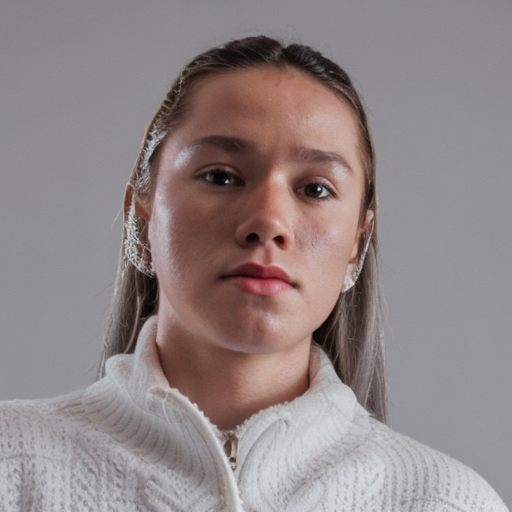} &
    \includegraphics[width=.14\linewidth]{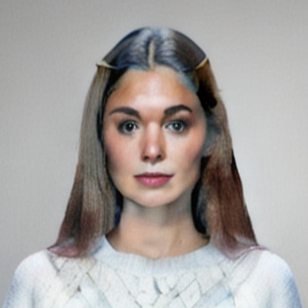} &
    \includegraphics[width=.14\linewidth]{images/frll/bonafide/082_03.jpg} \\

\end{tabular}
\endgroup
\caption{Comparison of morphed images generated with the four configurations tested in the ablation study vs. the default one.}
\label{fig:qualitative-ablation}
\end{figure*}

\subsection{Ablation Study}
\label{subsec:ablation}

To verify the soundness of the various design decisions we made in StableMorph, we performed an ablation study to evaluate how different configurations could affect the results. In the default StableMorph configuration, we propose merging the weights of the two LoRAs (one for each of the subjects) and plugging in the identity information of both subjects extracted from 10 images per subject. Therefore, we conduct the ablation study with four different configurations: (A) using LoRAs with only three images per subject to extract the identity; (B) using LoRAs with only five images per subject to extract the identity; (C) using the identity information with 10 images per subject but without using LoRAs; (D) using LoRAs only but without any identity information. We found the quantitative and qualitative results of the first two configurations (A and B) to be rather similar to the ones in the default StableMorph configuration (10 images). However, the generated images frequently contained some noise artifacts on the face and other areas and were generally less rich in details. This means that one can still use StableMorph to generate a morphed image if they have fewer images per subject, and they will still obtain the same crisp, sharp, high-quality morphed images, but it will probably require running the generation process more than once and experimenting with different seeds and numbers of inference steps. On the other hand, the results of configurations C and D were unsatisfactory. In configuration C (without LoRAs), the generated images drift too far away from the images of the two subjects. In configuration D (without identity information), the generated images almost completely ignore the subjects and revert to some average images internalized in the model. Figure \ref{fig:qualitative-ablation} shows a comparison of images produced by each of the four configurations as well as the default StableMorph configuration. The results solidify the choices made in StableMorph to use both merging LoRAs weights and increasing the number of images used to extract the identity information when possible.

\section{Discussion}
\label{sec:discussion}

\begin{figure}[t]
\centering
\begingroup
\setlength{\tabcolsep}{1.3pt} % Default value: 6pt
\begin{tabular}{ccc}
    
    noise on face &
    accessory &
    confusing gender \\
    
    \includegraphics[width=.32\linewidth]{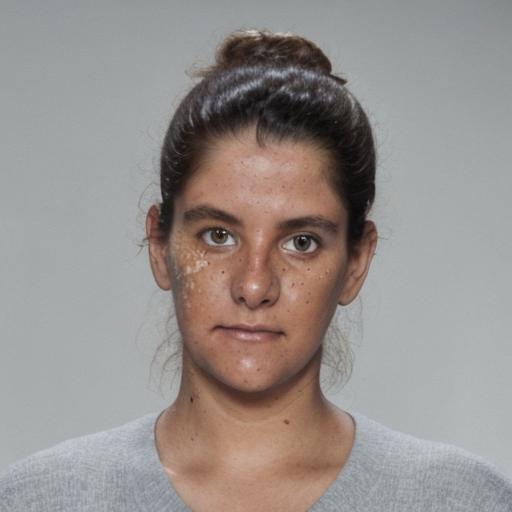} &
    \includegraphics[width=.32\linewidth]{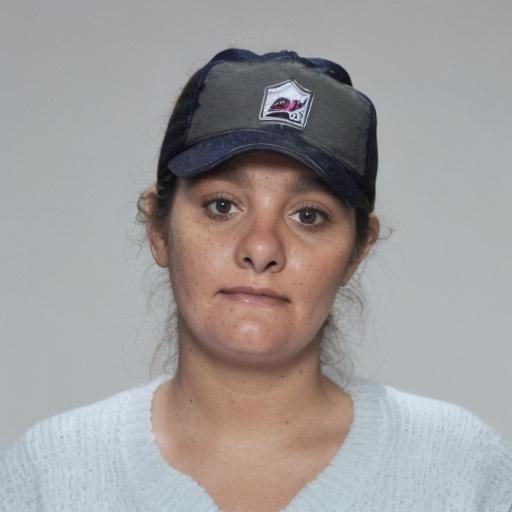} &
    \includegraphics[width=.32\linewidth]{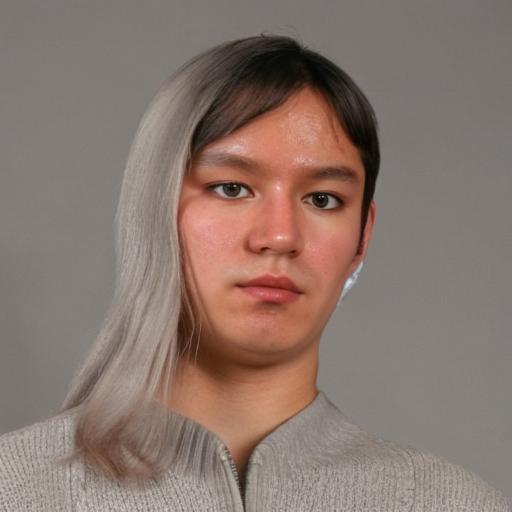} \\
    
    \includegraphics[width=.32\linewidth]{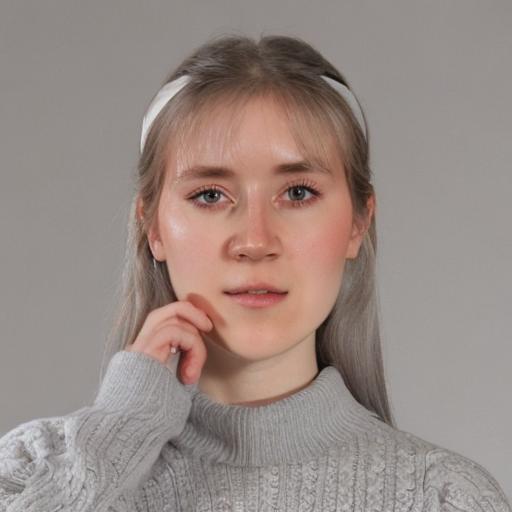} &
    \includegraphics[width=.32\linewidth]{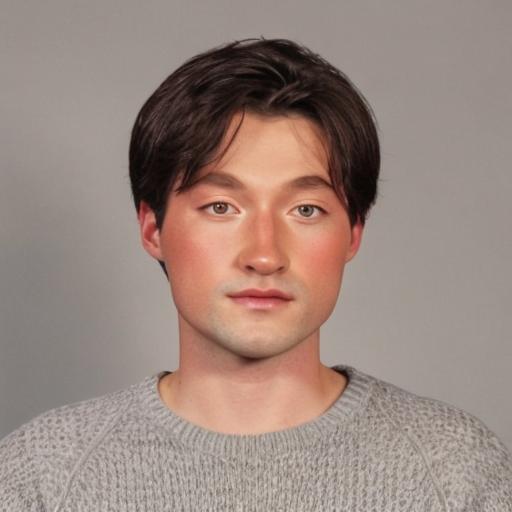} &
    \includegraphics[width=.32\linewidth]{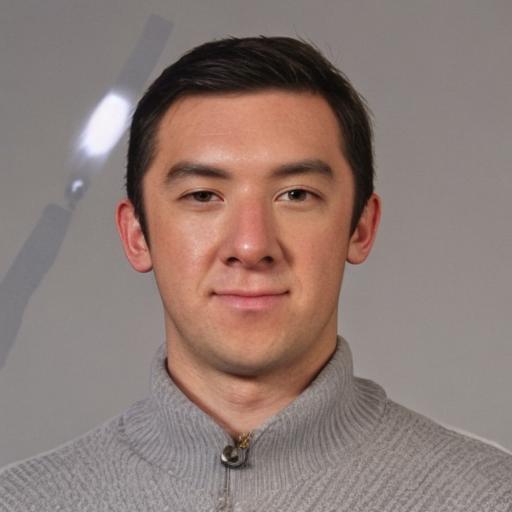} \\
    
    hand &
    painting effect & 
    bg. artifact \\
    
\end{tabular}
\endgroup

\caption{Samples of various defects encountered in some of the generated images.}
\label{fig:discussion-defects}
\end{figure}

\begin{figure}[t]
\centering

\begingroup
\setlength{\tabcolsep}{1.3pt} % Default value: 6pt
\begin{tabular}{ccc}
    
    Subject 1 &
    blue eyes &
    Subject 2 \\
    
    \includegraphics[width=.32\linewidth]{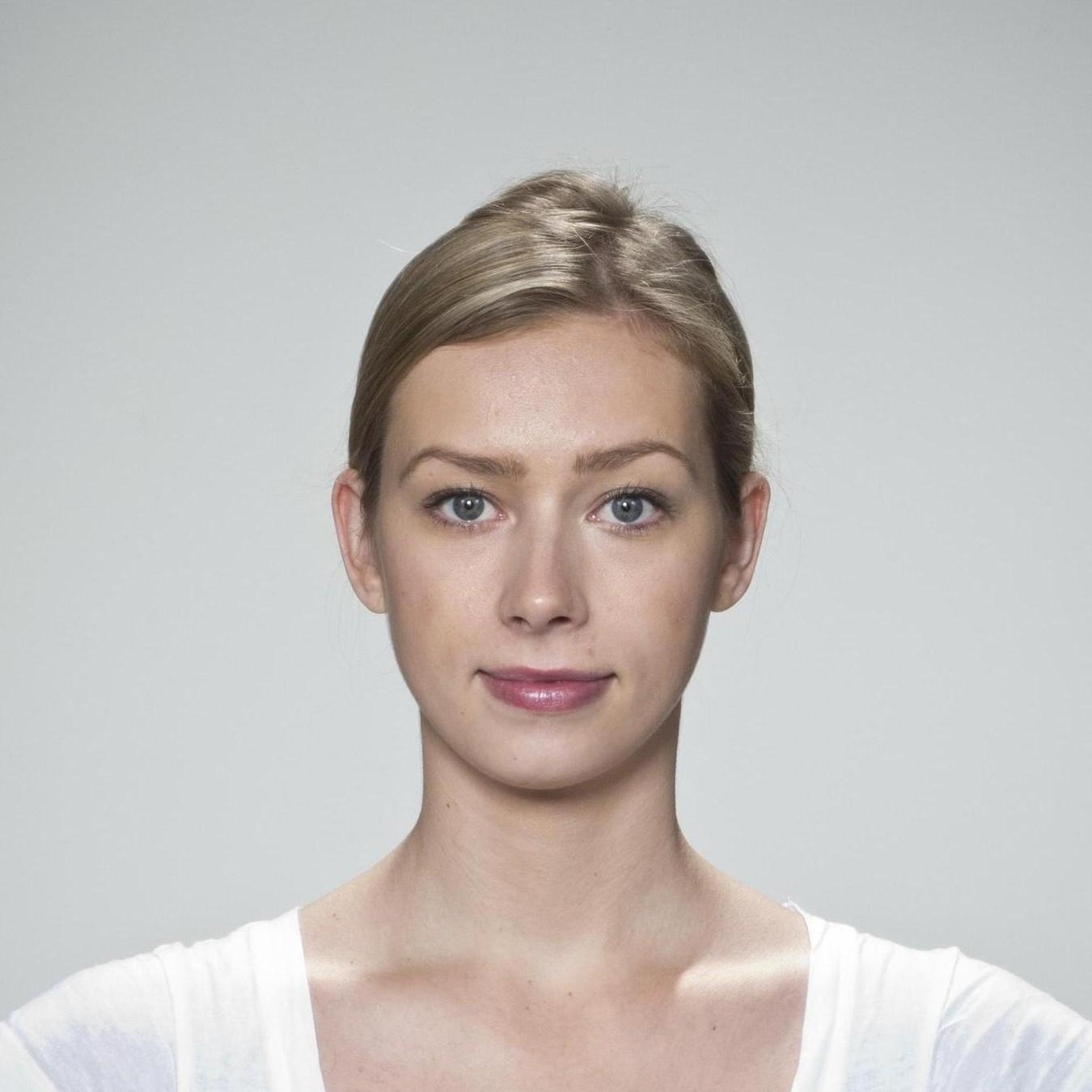} &
    \includegraphics[width=.32\linewidth]{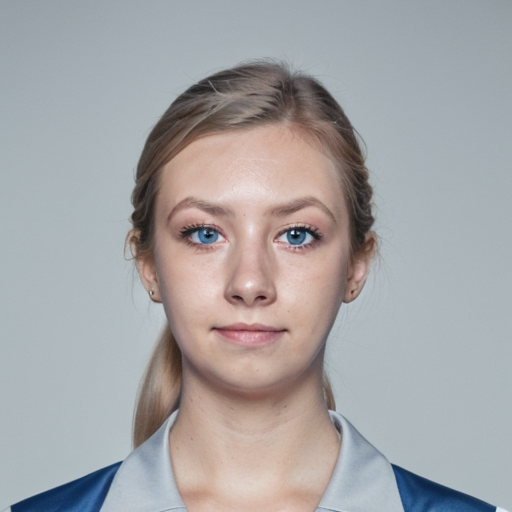} &
    \includegraphics[width=.32\linewidth]{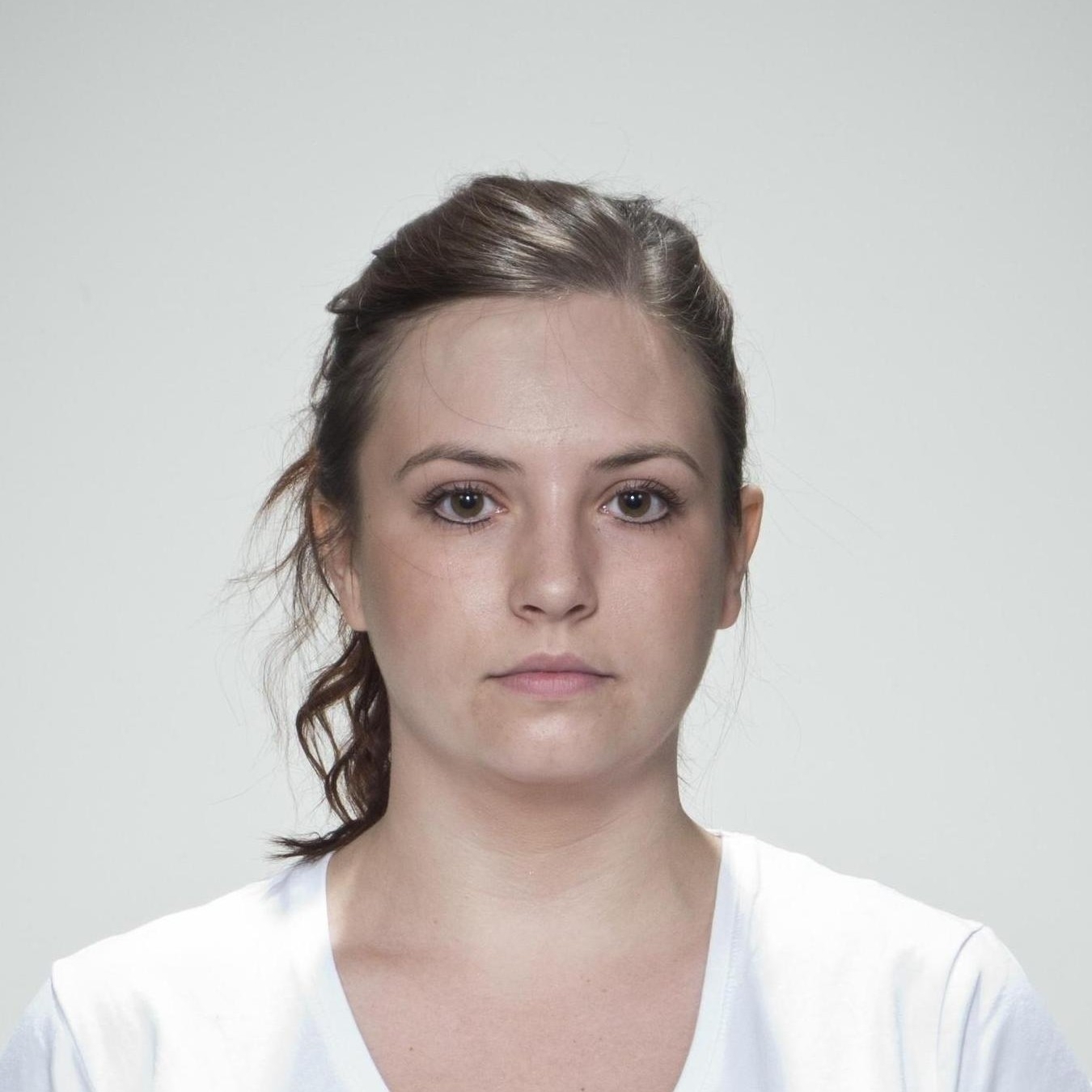} \\
    
    \includegraphics[width=.32\linewidth]{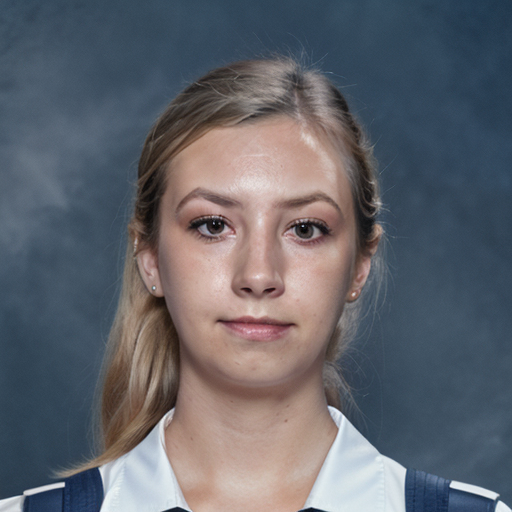} &
    \includegraphics[width=.32\linewidth]{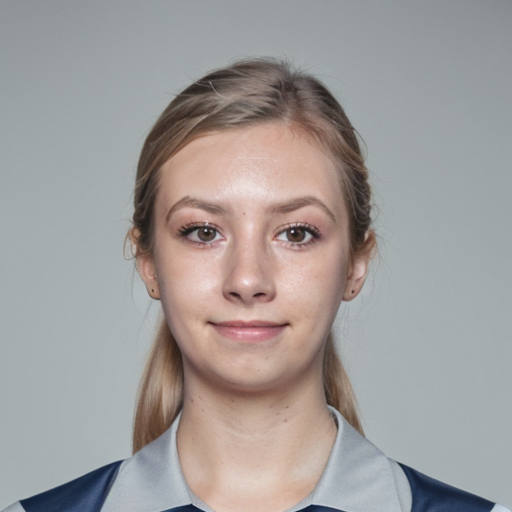} &
    \includegraphics[width=.32\linewidth]{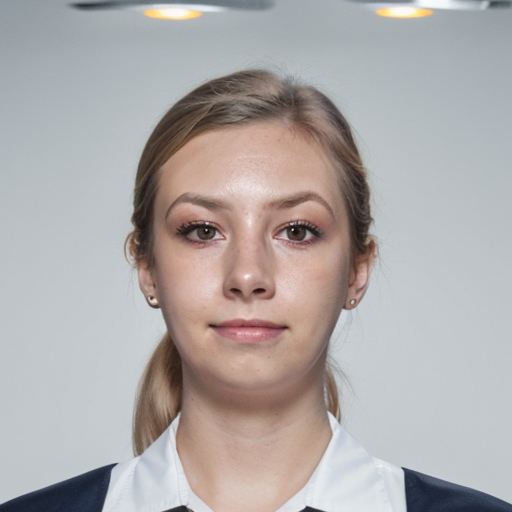} \\

    background &
    smiling & 
    makeup
\end{tabular}
\endgroup

\caption{Examples of the ability to control various attributes of the generated morphed image.}
\label{fig:discussion-controlling}
\end{figure}

Despite the fact that StableMorph generates morphed images with superior quality, it does occasionally produce images with undesired defects. These defects could be noise on some parts of the face, added accessories, confusing genders, a hand in front of the face, a painting effect, or artifacts in the background. Figure \ref{fig:discussion-defects} shows examples of these defects. We believe this happens for two reasons. First is the inherent randomness of the diffusion process, which produces different results given different initial noise. Second, given that SD is a text-to-image diffusion model, i.e., the generation process is also affected by the textual prompt, and for the mass generation of morphed images in our experiments, we cannot in practice tune this textual prompt and fit it to each pair of images. So, we use a single generic prompt that we found to generate the best possible morphed images on different of input images, without, for example, specifying whether the generated image should be a male or a female. But this also gives StableMorph a unique and interesting feature, which allows one to control the generation process using the textual prompt and further tune and adjust the morphed image. Besides, given the randomness of the generation process, several morphed images can be generated, and the best one can be selected. This allows one to tailor the generated morphs to the intended use, which could be very helpful in a real-world attack scenario. Figure \ref{fig:discussion-controlling} shows a sample morphed image where the textual prompt is used to manipulate some attributes of the generated image.

\section{Conclusion}
\label{sec:conclusion}

In this work, we introduced StableMorph, a novel approach for generating high-quality, realistic face morphs using \textit{Stable Diffusion}. In contrast to existing methods that often produce low-quality or artifact-prone images, StableMorph creates sharp, natural-looking morphs that closely resemble genuine photos—making them harder to detect by both humans and automated systems. By significantly improving the visual quality and realism of morphed images while maintaining strong attack potential, StableMorph presents a more challenging benchmark for evaluating the robustness of face recognition systems. This capability is especially valuable for developing and testing next-generation morphing attack detection (MAD) methods, pushing the field toward more resilient biometric security solutions.

\section*{Acknowledgement}
\label{sec:acknowledgment}

This research has received funding from the European Union’s Horizon Europe research and innovation programme under grant agreement No 101121280 (EINSTEIN). Views and opinions expressed are however those of the author(s) only and do not necessarily reflect the views of the EU/Executive Agency. Neither the EU or the granting authority can be held responsible for them.

\clearpage

{\small
\bibliographystyle{ieee}
\bibliography{egbib}

@String(CVPR= {IEEE Conf. Comput. Vis. Pattern Recog.})

@String(CVPR  = {CVPR})

@inproceedings{ou2021sdd,
  title={SDD-FIQA: Unsupervised face image quality assessment with similarity distribution distance},
  author={Ou, Fu-Zhao and Chen, Xingyu and Zhang, Ruixin and Huang, Yuge and Li, Shaoxin and Li, Jilin and Li, Yong and Cao, Liujuan and Wang, Yuan-Gen},
  booktitle={Proceedings of the IEEE/CVF conference on computer vision and pattern recognition},
  pages={7670--7679},
  year={2021}
}

@article{shah2023ZipLoRA,
  title={ZipLoRA: Any Subject in Any Style by Effectively Merging LoRAs},
  author={Shah, Viraj and Ruiz, Nataniel and Cole, Forrester and Lu, Erika and Lazebnik, Svetlana and Li, Yuanzhen and Jampani, Varun},
  booktitle={arXiv preprint arxiv:2311.13600},
  year={2023}
}

@InProceedings{Zhang_2024_CVPR,
    author    = {Zhang, Kaiwen and Zhou, Yifan and Xu, Xudong and Dai, Bo and Pan, Xingang},
    title     = {DiffMorpher: Unleashing the Capability of Diffusion Models for Image Morphing},
    booktitle = {Proceedings of the IEEE/CVF Conference on Computer Vision and Pattern Recognition (CVPR)},
    month     = {June},
    year      = {2024},
    pages     = {7912-7921}
}

@inproceedings{schardong2024neural,
  title={Neural implicit morphing of face images},
  author={Schardong, Guilherme and Novello, Tiago and Paz, Hallison and Medvedev, Iurii and da Silva, Vin{\'\i}cius and Velho, Luiz and Gon{\c{c}}alves, Nuno},
  booktitle={Proceedings of the IEEE/CVF Conference on Computer Vision and Pattern Recognition},
  pages={7321--7330},
  year={2024}
}

@InProceedings{grimmer2024ladimo,
    author = {Grimmer, Marcel and Busch, Christoph},
    title = {LADIMO: Face Morph Generation through Biometric Template Inversion with Latent Diffusion},
    booktitle = {Proceedings of International Joint Conference on Biometrics (IJCB)},
    month     = {September},
    year      = {2024},
}

@article{zhong2021sface,
  title={Sface: Sigmoid-constrained hypersphere loss for robust face recognition},
  author={Zhong, Yaoyao and Deng, Weihong and Hu, Jiani and Zhao, Dongyue and Li, Xian and Wen, Dongchao},
  journal={IEEE Transactions on Image Processing},
  volume={30},
  pages={2587--2598},
  year={2021},
  publisher={IEEE}
}

@inproceedings{rombach2022high,
  title={High-resolution image synthesis with latent diffusion models},
  author={Rombach, Robin and Blattmann, Andreas and Lorenz, Dominik and Esser, Patrick and Ommer, Bj{\"o}rn},
  booktitle={Proceedings of the IEEE/CVF conference on computer vision and pattern recognition},
  pages={10684--10695},
  year={2022}
}

@article{Fu2022quality,
author = {Fu, Biying and Damer, Naser},
title = {Face morphing attacks and face image quality: The effect of morphing and the unsupervised attack detection by quality},
journal = {IET Biometrics},
volume = {11},
number = {5},
pages = {359-382},
doi = {https://doi.org/10.1049/bme2.12094},
url = {https://ietresearch.onlinelibrary.wiley.com/doi/abs/10.1049/bme2.12094},
eprint = {https://ietresearch.onlinelibrary.wiley.com/doi/pdf/10.1049/bme2.12094},
year = {2022}
}

@INPROCEEDINGS{Makrushin2018Mitigating,
  author={Makrushin, Andrey and Wolf, Andreas},
  booktitle={2018 26th European Signal Processing Conference (EUSIPCO)}, 
  title={An Overview of Recent Advances in Assessing and Mitigating the Face Morphing Attack}, 
  year={2018},
  volume={},
  number={},
  pages={1017-1021},
  doi={10.23919/EUSIPCO.2018.8553599}}

@inproceedings{Kraetzer2017Media-Forensics,
author = {Kraetzer, Christian and Makrushin, Andrey and Neubert, Tom and Hildebrandt, Mario and Dittmann, Jana},
title = {Modeling Attacks on Photo-ID Documents and Applying Media Forensics for the Detection of Facial Morphing},
year = {2017},
isbn = {9781450350617},
publisher = {Association for Computing Machinery},
address = {New York, NY, USA},
url = {https://doi.org/10.1145/3082031.3083244},
doi = {10.1145/3082031.3083244},
booktitle = {Proceedings of the 5th ACM Workshop on Information Hiding and Multimedia Security},
pages = {21–32},
numpages = {12},
location = {Philadelphia, Pennsylvania, USA},
series = {IH and MMSec '17}
}

@inproceedings{kim2022adaface,
  title={AdaFace: Quality Adaptive Margin for Face Recognition},
  author={Kim, Minchul and Jain, Anil K and Liu, Xiaoming},
  booktitle={Proceedings of the IEEE/CVF Conference on Computer Vision and Pattern Recognition},
  year={2022}
}

@inproceedings{parmar2022aliased,
  title={On aliased resizing and surprising subtleties in gan evaluation},
  author={Parmar, Gaurav and Zhang, Richard and Zhu, Jun-Yan},
  booktitle={Proceedings of the IEEE/CVF Conference on Computer Vision and Pattern Recognition},
  pages={11410--11420},
  year={2022}
}

@article{heusel2017gans,
  title={Gans trained by a two time-scale update rule converge to a local nash equilibrium},
  author={Heusel, Martin and Ramsauer, Hubert and Unterthiner, Thomas and Nessler, Bernhard and Hochreiter, Sepp},
  journal={Advances in neural information processing systems},
  volume={30},
  year={2017}
}

@article{merkle2022state,
  title={State of the art of quality assessment of facial images},
  author={Merkle, Johannes and Rathgeb, Christian and Tams, Benjamin and Lou, Dhay-Parn and D{\"o}rsch, Andr{\'e} and Drozdowski, Pawel},
  journal={arXiv preprint arXiv:2211.08030},
  year={2022}
}

@inproceedings{lao2022attentions,
  title={Attentions help cnns see better: Attention-based hybrid image quality assessment network},
  author={Lao, Shanshan and Gong, Yuan and Shi, Shuwei and Yang, Sidi and Wu, Tianhe and Wang, Jiahao and Xia, Weihao and Yang, Yujiu},
  booktitle={Proceedings of the IEEE/CVF conference on computer vision and pattern recognition},
  pages={1140--1149},
  year={2022}
}

@inproceedings{cheon2021perceptual,
  title={Perceptual image quality assessment with transformers},
  author={Cheon, Manri and Yoon, Sung-Jun and Kang, Byungyeon and Lee, Junwoo},
  booktitle={Proceedings of the IEEE/CVF conference on computer vision and pattern recognition},
  pages={433--442},
  year={2021}
}

@article{mittal2012no,
  title={No-reference image quality assessment in the spatial domain},
  author={Mittal, Anish and Moorthy, Anush Krishna and Bovik, Alan Conrad},
  journal={IEEE Transactions on image processing},
  volume={21},
  number={12},
  pages={4695--4708},
  year={2012},
  publisher={IEEE}
}

@inproceedings{ying2020patches,
  title={From patches to pictures (PaQ-2-PiQ): Mapping the perceptual space of picture quality},
  author={Ying, Zhenqiang and Niu, Haoran and Gupta, Praful and Mahajan, Dhruv and Ghadiyaram, Deepti and Bovik, Alan},
  booktitle={Proceedings of the IEEE/CVF conference on computer vision and pattern recognition},
  pages={3575--3585},
  year={2020}
}

@inproceedings{yang2022maniqa,
  title={Maniqa: Multi-dimension attention network for no-reference image quality assessment},
  author={Yang, Sidi and Wu, Tianhe and Shi, Shuwei and Lao, Shanshan and Gong, Yuan and Cao, Mingdeng and Wang, Jiahao and Yang, Yujiu},
  booktitle={Proceedings of the IEEE/CVF Conference on Computer Vision and Pattern Recognition},
  pages={1191--1200},
  year={2022}
}

@inproceedings{zhang2023adding,
  title={Adding conditional control to text-to-image diffusion models},
  author={Zhang, Lvmin and Rao, Anyi and Agrawala, Maneesh},
  booktitle={Proceedings of the IEEE/CVF International Conference on Computer Vision},
  pages={3836--3847},
  year={2023}
}

@article{ye2023ip-adapter,
  title={Ip-adapter: Text compatible image prompt adapter for text-to-image diffusion models},
  author={Ye, Hu and Zhang, Jun and Liu, Sibo and Han, Xiao and Yang, Wei},
  journal={arXiv preprint arXiv:2308.06721},
  year={2023}
}

@inproceedings{hu2022lora,
title={Lo{RA}: Low-Rank Adaptation of Large Language Models},
author={Edward J Hu and yelong shen and Phillip Wallis and Zeyuan Allen-Zhu and Yuanzhi Li and Shean Wang and Lu Wang and Weizhu Chen},
booktitle={International Conference on Learning Representations},
year={2022},
url={https://openreview.net/forum?id=nZeVKeeFYf9}
}

@article{dhariwal2021diffusion,
  title={Diffusion models beat gans on image synthesis},
  author={Dhariwal, Prafulla and Nichol, Alexander},
  journal={Advances in neural information processing systems},
  volume={34},
  pages={8780--8794},
  year={2021}
}

@article{ho2020denoising,
  title={Denoising diffusion probabilistic models},
  author={Ho, Jonathan and Jain, Ajay and Abbeel, Pieter},
  journal={Advances in neural information processing systems},
  volume={33},
  pages={6840--6851},
  year={2020}
}

@misc{sohldickstein2015deep,
      title={Deep Unsupervised Learning using Nonequilibrium Thermodynamics}, 
      author={Jascha Sohl-Dickstein and Eric A. Weiss and Niru Maheswaranathan and Surya Ganguli},
      year={2015},
      eprint={1503.03585},
      archivePrefix={arXiv},
      primaryClass={id='cs.LG' full_name='Machine Learning' is_active=True alt_name=None in_archive='cs' is_general=False description='Papers on all aspects of machine learning research (supervised, unsupervised, reinforcement learning, bandit problems, and so on) including also robustness, explanation, fairness, and methodology. cs.LG is also an appropriate primary category for applications of machine learning methods.'}
}

@inproceedings{esser2024scaling,
  title={Scaling rectified flow transformers for high-resolution image synthesis},
  author={Esser, Patrick and Kulal, Sumith and Blattmann, Andreas and Entezari, Rahim and M{\"u}ller, Jonas and Saini, Harry and Levi, Yam and Lorenz, Dominik and Sauer, Axel and Boesel, Frederic and others},
  booktitle={Forty-first International Conference on Machine Learning},
  year={2024}
}

@misc{rombach2022highresolution,
      title={High-Resolution Image Synthesis with Latent Diffusion Models}, 
      author={Robin Rombach and Andreas Blattmann and Dominik Lorenz and Patrick Esser and Björn Ommer},
      year={2022},
      eprint={2112.10752},
      archivePrefix={arXiv},
      primaryClass={id='cs.CV' full_name='Computer Vision and Pattern Recognition' is_active=True alt_name=None in_archive='cs' is_general=False description='Covers image processing, computer vision, pattern recognition, and scene understanding. Roughly includes material in ACM Subject Classes I.2.10, I.4, and I.5.'}
}

@ARTICLE{Cao2024GDM,
  author={Cao, Hanqun and Tan, Cheng and Gao, Zhangyang and Xu, Yilun and Chen, Guangyong and Heng, Pheng-Ann and Li, Stan Z.},
  journal={IEEE Transactions on Knowledge and Data Engineering}, 
  title={A Survey on Generative Diffusion Models}, 
  year={2024},
  volume={36},
  number={7},
  pages={2814-2830},
  doi={10.1109/TKDE.2024.3361474}}

@article{Yang2023DMA,
author = {Yang, Ling and Zhang, Zhilong and Song, Yang and Hong, Shenda and Xu, Runsheng and Zhao, Yue and Zhang, Wentao and Cui, Bin and Yang, Ming-Hsuan},
title = {Diffusion Models: A Comprehensive Survey of Methods and Applications},
year = {2023},
issue_date = {April 2024},
publisher = {Association for Computing Machinery},
address = {New York, NY, USA},
volume = {56},
number = {4},
issn = {0360-0300},
url = {https://doi.org/10.1145/3626235},
doi = {10.1145/3626235},
journal = {ACM Comput. Surv.},
month = {nov},
articleno = {105},
numpages = {39}
}

@ARTICLE{Croitoru2023DMV,
  author={Croitoru, Florinel-Alin and Hondru, Vlad and Ionescu, Radu Tudor and Shah, Mubarak},
  journal={IEEE Transactions on Pattern Analysis and Machine Intelligence}, 
  title={Diffusion Models in Vision: A Survey}, 
  year={2023},
  volume={45},
  number={9},
  pages={10850-10869},
  doi={10.1109/TPAMI.2023.3261988}}

@article{Liao2014MorphingSSIM,
author = {Liao, Jing and Lima, Rodolfo S. and Nehab, Diego and Hoppe, Hugues and Sander, Pedro V. and Yu, Jinhui},
title = {Automating Image Morphing Using Structural Similarity on a Halfway Domain},
year = {2014},
issue_date = {August 2014},
publisher = {Association for Computing Machinery},
address = {New York, NY, USA},
volume = {33},
number = {5},
issn = {0730-0301},
url = {https://doi.org/10.1145/2629494},
doi = {10.1145/2629494},
journal = {ACM Trans. Graph.},
month = {sep},
articleno = {168},
numpages = {12}
}

@INPROCEEDINGS{Damer2023MorDiff,
  author={Damer, Naser and Fang, Meiling and Siebke, Patrick and Kolf, Jan Niklas and Huber, Marco and Boutros, Fadi},
  booktitle={2023 11th International Workshop on Biometrics and Forensics (IWBF)}, 
  title={MorDIFF: Recognition Vulnerability and Attack Detectability of Face Morphing Attacks Created by Diffusion Autoencoders}, 
  year={2023},
  volume={},
  number={},
  pages={1-6},
  doi={10.1109/IWBF57495.2023.10157869}}

@inproceedings{venkatesh2020can,
  title={Can GAN generated morphs threaten face recognition systems equally as landmark based morphs?-vulnerability and detection},
  author={Venkatesh, Sushma and Zhang, Haoyu and Ramachandra, Raghavendra and Raja, Kiran and Damer, Naser and Busch, Christoph},
  booktitle={2020 8th International Workshop on Biometrics and Forensics (IWBF)},
  pages={1--6},
  year={2020},
  organization={IEEE}
}

@article{DeBruine2021,
author = "Lisa DeBruine and Benedict Jones",
title = "{Face Research Lab London Set}",
year = "2021",
month = "4",
url = "https://figshare.com/articles/dataset/Face_Research_Lab_London_Set/5047666",
doi = "10.6084/m9.figshare.5047666.v5"
}

@inproceedings{Boutros-CR-FIQA-CVPR-2023,
 Author = {F. Boutros and M. Fang and M. Klemt and B. Fu and N. Damer},
 Booktitle = {Conf. on Computer Vision and Pattern Recognition {(CVPR)}},
 Organization = {IEEE},
 Pages = {5836--5845},
 Title = {{CR-FIQA}: Face Image Quality Assessment by Learning Sample Relative Classifiability},
 Year = {2023}
}

@inproceedings{Damer-MorGAN-BTAS-2018,
 Author = {N. Damer and Y. Wainakh and V. Boller and S. {von den Berken} and P. Terh{\"o}rst and A. Braun and A. Kuijper},
 Booktitle = {Proc. of the 9th {IEEE} Intl. Conf. on Biometrics: Theory, Applications, and Systems ({BTAS})},
 Publisher = {IEEE},
 Title = {{MorGAN}: Recognition Vulnerability and Attack Detectability of Face Morphing Attacks Created by Generative Adversarial Network},
 Year = {2018}
}

@inproceedings{Deng-ArcFace-IEEE-CVPR-2019,
 Author = {J. Deng and J. Guo and S. Zafeiriou},
 Booktitle = {Conf. on Computer Vision and Pattern Recognition ({CVPR})},
 Month = {June},
 Title = {{ArcFace}: Additive Angular Margin Loss for Deep Face Recognition},
 Year = {2019}
}

@inproceedings{Ferrara-MAP-IWBF-2022,
 Author = {M. Ferrara and A. Franco and D. Maltoni and C. Busch},
 Booktitle = {10th Intl. Workshop on Biometrics and Forensics ({IWBF})},
 Month = {April},
 Title = {Morphing Attack Potential},
 Year = {2022}
}

@inproceedings{Ferrara-TextureBlendingAndShapeWarpingInFaceMorphing-IEEE-BIOSIG-2019,
 Author = {M. Ferrara and A. Franco and D. Maltoni},
 Booktitle = {Intl. Conf. of the Biometrics Special Interest Group ({BIOSIG})},
 Month = {September},
 Publisher = {IEEE},
 Title = {Decoupling texture blending and shape warping in face morphing},
 Year = {2019}
}

@inproceedings{Ferrara-TheMagicPassport-IJCB-2014,
 Author = {M. Ferrara and A. Franco and D. Maltoni},
 Booktitle = {2014 {IEEE} Intl. Joint Conf. on Biometrics ({IJCB})},
 Month = {September},
 Pages = {1--7},
 Title = {The magic passport},
 Year = {2014}
}

@techreport{ICAO-PortraitQuality-TR-2018,
 Author = {{ISO/IEC JTC1 SC17 WG3}},
 Institution = {{International Civil Aviation Organization}},
 Month = {April},
 Pages = {85},
 Title = {{Portrait Quality} - {Reference Facial Images for MRTD}},
 Year = {2018}
}

@manual{ISO-IEC-20059,
 Author = {{ISO/IEC JTC1 SC37 Biometrics}},
 Organization = {International Organization for Standardization},
 Title = {{ISO/IEC} {CD2} 20059. Information Technology -- Methodologies to evaluate the resistance of biometric recognition systems to morphing attacks},
 Year = {2024}
}

@manual{ISO-IEC-29794-5-DIS-FaceQuality-240129,
 Author = {{ISO/IEC JTC1 SC37 Biometrics}},
 Organization = {International Organization for Standardization},
 Title = {{ISO/IEC} {DIS} 29794-5 Information Technology - Biometric Sample Quality - Part 5: Face Image Data},
 Year = {2024}
}

@inproceedings{Makrushin-FaultlessMorphs-ICVICGTA-2017,
 Author = {A. Makrushin and T. Neubert and J. Dittmann},
 Booktitle = {Proc. of the 12th Intl. Joint Conf. on Computer Vision, Imaging and Computer Graphics Theory and Applications},
 Publisher = {{SCITEPRESS} - Science and Technology Publications},
 Title = {Automatic Generation and Detection of Visually Faultless Facial Morphs},
 Year = {2017}
}

@inproceedings{Meng-FRwithFQA-MagFace-CVPR-2021,
 Author = {Q. Meng and S. Zhao and Z. Huang and F. Zhou},
 Booktitle = {2021 {IEEE}/{CVF} Conf. on Computer Vision and Pattern Recognition ({CVPR})},
 Link = {http://arxiv.org/abs/2103.06627},
 Month = {March},
 Title = {{{MagFace}}: A Universal Representation for Face Recognition and Quality Assessment},
 Year = {2021}
}

@inproceedings{Phillips-OverviewFaceRecognitionGrandChallengeFRGC-CVPR-2005,
 Author = {J. Phillips and P. Flynn and T. Scruggs and K. Bowyer and J. Chang and others},
 Booktitle = {Conf. on {{Computer Vision}} and {{Pattern Recognition}} ({{CVPR}})},
 Issn = {1063-6919},
 Month = {June},
 Pages = {947--954},
 Publisher = {IEEE},
 Title = {Overview of the {{Face Recognition Grand Challenge}}},
 Volume = {1},
 Year = {2005}
}

@inproceedings{Raghavendra-DetectingMorphedFace-BTAS-2016,
 Author = {R. Raghavendra and K. Raja and C. Busch},
 Booktitle = {2016 {IEEE} 8th Intl. Conf. on Biometrics: Theory, Applications and Systems ({BTAS})},
 Month = {September},
 Organization = {8th {IEEE} Intl. Conf. on Biometrics: Theory, Applications and Systems (BTAS-2016)},
 Publisher = {IEEE},
 Title = {Detecting morphed face images},
 Year = {2016}
}

@inproceedings{Scherhag-MorphingAttacks-MorphingTechniques-BIOSIG-2017,
 Author = {U. Scherhag and A. Nautsch and C. Rathgeb and M. Gomez-Barrero and R. N. J. Veldhuis and L. Spreeuwers and M. Schils and D. Maltoni and P. Grother and S. Marcel and R. Breithaupt and R. Raghavendra and C. Busch},
 Booktitle = {Intl. Conf. of the Biometrics Special Interest Group {BIOSIG} 2017},
 Pages = {1--7},
 Title = {Biometric Systems under Morphing Attacks: Assessment of Morphing Techniques and Vulnerability Reporting},
 Year = {2017}
}

@article{Scherhag-MorphingAttacks-Survey-IEEEAccess-2019,
 Author = {U. Scherhag and C. Rathgeb and J. Merkle and R. Breithaupt and C. Busch},
 Journal = {IEEEAccess},
 Title = {Face Recognition Systems under Morphing Attacks: A Survey},
 Year = {2019}
}

@inproceedings{Scherhag-MorphingDetection-ICBEA-2018,
 Author = {U. Scherhag and C. Rathgeb and C. Busch},
 Booktitle = {Intl. Conf. on Biometric Engineering and Applications 2018 ({ICBEA})},
 Pages = {1--7},
 Title = {Morph detection from single face images: a multi-algorithm fusion approach},
 Year = {2018}
}

@article{Scherhag-PRNU-TBIOM-2019,
 Author = {U. Scherhag and L. Debiasi and C. Rathgeb and C. Busch and A. Uhl},
 Journal = {Trans. on Biometrics, Behavior, and Identity Science ({TBIOM})},
 Title = {Detection of Face Morphing Attacks based on {PRNU} Analysis},
 Year = {2019}
}

@article{Schlett-FIQA-LiteratureSurvey-CSUR-2021,
 Author = {T. Schlett and C. Rathgeb and O. Henniger and J. Galbally and J. Fierrez and C. Busch},
 Issn = {0360-0300},
 Journal = {{ACM} Computing Surveys ({CSUR})},
 Month = {December},
 Publisher = {ACM},
 Title = {Face Image Quality Assessment: A Literature Survey},
 Year = {2021}
}

@article{Venkatesh-FaceMorphingAttackGenerationAndDetection-TTS-2021,
 Author = {S. Venkatesh and R. Raghavendra and K. Raja and C. Busch},
 Journal = {{IEEE} Trans. on Technology and Society},
 Month = {September},
 Title = {Face Morphing Attack Generation \& Detection: A Comprehensive Survey},
 Year = {2021}
}

@article{Zhang-MIPGAN-MorphingAttacks-IEEE-2021,
 Author = {H. Zhang and S. Venkatesh and R. Raghavendra and K. Raja and N. Damer and C. Busch},
 Journal = {{IEEE} Trans. on Biometrics, Behaviour and Identity},
 Month = {September},
 Title = {{MIPGAN} -- Generating Strong and High Quality Morphing Attacks Using Identity Prior Driven {GAN}},
 Year = {2021}
}
}

\end{document}